\documentclass{article}

\usepackage{graphicx} 
\usepackage{subcaption}
\usepackage{float}
\usepackage{booktabs}
\usepackage{makecell}
\usepackage{amsmath}
\usepackage{xcolor}
\usepackage{longtable}
\usepackage{multirow}
\usepackage[backend=biber,style=apa]{biblatex}
\usepackage{tikz}
\usetikzlibrary{positioning, arrows.meta, calc}
\usepackage{authblk}

\title{
    \textbf{A Data-Driven Approach to State Construction in Markov Models}
}

\author[1,*]{Linde Van Gestel}
\author[1]{Marie-Anne Guerry}
\author[2]{Evy Rombaut}
\affil[1]{
    Department of Business Technology and Operations, Data Analytics Laboratory,
    Vrije Universiteit Brussel, Pleinlaan 2, 1050 Brussels, Belgium
}
\affil[2]{
    Mobility and Automotive Technology,
    Vrije Universiteit Brussel, Pleinlaan 2, 1050 Brussels, Belgium
}
\affil[*]{
    Corresponding author: \texttt{linde.van.gestel@vub.be}
}
\date{August 2026}

\begin{document}

\maketitle

\section*{Abstract}
A Markov chain is a widely used stochastic process modelling random events over time. These models are built on subsets of the entire dataset, referred to as states, which are considered to be homogeneous regarding transition probabilities. However, the creation of these states is often disregarded or based on prior assumption, potentially violating the homogeneity requirement and thus decreasing the validity and predictive power of the model. In order to fill this gap, this paper combines supervised feature selection with unsupervised learning techniques for data-driven state construction. Density-based clustering, spectral clustering, and Kohonen self-organizing maps are examined for their ability to identify latent groups without prior assumptions. The contribution of this study is twofold. First, the paper presents a methodological framework for state construction incorporating suitable unsupervised learning techniques, with appropriate measures both for classification performance and Markov model accuracy. Secondly, the framework is tested on an application, resulting in a comparative analysis showing that spectral clustering and Kohonen self-organizing maps are best at capturing inherent structure. These results serve as a cornerstone in providing theoretical and methodological guidance for improving state definition in applied Markov modelling.
\vspace{0.5cm}

\textbf{Keywords}: Markov model, States, Homogeneity, (Un)Supervised learning, Clustering

\newpage
\section{Introduction}
A Markov chain is a stochastic process modelling random events over time. Applications are widespread, ranging from the analysis of genetic algorithms \parencite{suzuki1995markov} in biology to manpower planning models \parencite{mcclean1991manpower}, as well as the measurement of economic trends such as income disparities \parencite{kerkouch2024dynamic} and share price movements \parencite{manasseh2022application}. 
\vspace{0.5cm}

These Markov models are constructed on subsets of the entire dataset considered to be homogeneous regarding transition probabilities, also referred to as states. However, the creation of these states is often disregarded or neglected resulting in specification error, indicating that the assumption of homogeneous states is not met, which decreases the validity and predictive power of the model \parencite{bartholomew1975errors}. 
\vspace{0.5cm}

In determining states of a Markov model, prior research focused mainly on supervised techniques, such as logistic regression \parencite{de2006modelling} and decision trees \parencite{rombaut2016turnover}. 
\vspace{0.5cm}

This paper focuses on a novel, three-step grouping methodology developed and evaluated within the context of discrete-time, time-homogeneous Markov chains. First, the approach employs supervised feature selection \parencite{genuer2015vsurf} combined with unsupervised learning. The unsupervised learning, executed in the second step, adds the strength of simultaneous consideration of all variables informative for transition behavior, allowing to identify groups that are inherently present in the data. Observations that have similar values for the relevant variables, will thus be assumed to portray equal transition behavior. This rationale can be supported by statistical decision theory, where predictions are made with one function for all observations with similar input factors, assessed through nearest-neighbour methods \parencite{hastie2009elements}. The resulting subgroups, derived from hierarchical density-based clustering \parencite{campello2013density}, spectral clustering \parencite{von2007tutorial} or Kohonen self-organizing maps \parencite{kohonen2002self}, are merged in the third and last step to account for subgroups characterized by similar transition probabilities. An illustration is provided based on a manpower database, in which promotion and wastage is modeled and compared to states obtained in prior research using decision trees.

\section{Literature review}
\subsection{Homogeneous Subgroups}
Markov processes describe stochastic events over time. All possible outcomes of the event under study, are aggregated in states in the state space, which can be finite or infinite. If the state space of a Markov process is finite or countable, it is presented as a Markov chain \parencite{norris1998markov}. Markov chains are longitudinal in the sense that distinct entities can move from one state to another over time with a certain transition probability. A fundamental characteristic of a Markov model is that the probability that an identity will be in a certain state in the next period only depends on its current state \parencite{bremaud2013markov}. Numerous variations of this basic principle exist, however we focus on discrete, time-homogeneous Markov chains with a finite state space. Discrete implies that events occur at set time intervals.  The number of states in which entities can be found is limited, as the state space is finite. The probabilities are considered to be time-homogeneous, indicating that the probability to transfer from state \(i\) to state \(j\) does not change over time \parencite{sericola2013markov}.
\vspace{0.5cm}

Markov models are able to describe the transition of entities over time. This transition is made through classes, which are inherent to the research problem. Within each class, subclasses tend to exist, also referred to as the states, in which entities have similar probabilities of moving through the process. Events occur when entities move from one state to another in the next time period. For example, \textcite{bakhtiari2026prediction} used Markov models to predict consumer credit card risk. The research problem itself divides the observations into two classes: consumers who are able to pay back their debt and the ones who are not. However, people who are currently not at risk, have different probabilities of defaulting anyway at a certain point in time based on underlying characteristics. This raises the importance of introducing new states in order to build an appropriate model.
\vspace{0.5cm}

As these transition probabilities are determined for groups of observations, the observations within a Markov model do not need to be studied individually. Often, they are grouped together because this decreases computational effort and because it is representable for real-life examples. When creating states of the dataset, a fundamental assumption for the validity of a Markov model is that these states are homogeneous. This means that all members within a state have similar transition probabilities \parencite{bremaud2013markov}. If this assumption is not satisfied, it decreases the accuracy and predictive power of a Markov model. The importance of handling this heterogeneity has been pointed out prior by \textcite{de2006modelling} in a manpower planning context, stating that personnel systems tend to be split up only on hierarchy, without considering additional separation within each level, possibly distorting forecasting results. Similar remarks are made by \textcite{ugwuowo2000modelling}, indicating that the construction of states has a significant influence on wastage probabilities within a firm and thus should be treated with care. 
Nevertheless, the formation of these states is often disregarded or simplified.
\vspace{0.5cm}

Little research exists on how to reach these states for accurate Markov chain modeling, however, some approaches have been introduced \parencite{de2006modelling, rombaut2016turnover}. \textcite{de2006modelling} proposed a splitting-up process of several stages applied in a manpower planning context, which lies at the root of our novel framework. First, it splits up the dataset according to the classes related to the research questions under study. The study aims to examine how people are distributed across personnel grades, splitting up the dataset for each grade present. Secondly, the splitting process is being continued until a desirable level of homogeneity is reached. Subgroups are defined based on characteristics that display significantly different levels of transition probabilities, based on multinomial logistic regression, within each previously created subgroup. The cut-off splitting values for the variables are calculated through hierarchical clustering, using within-groups linkage resulting in binary splits. This brings us to the third and following stages where new transitions are introduced and homogeneity is investigated anew, only with respect to new introduced flows from a previous stage. This division goes on until an acceptable level of homogeneity is reached for all subgroups. During this process, one has to take into account the trade-off problem between homogeneity and the number of members in each group. Smaller groups increase homogeneity but explanation for the differences might not be reliable, as the maximum likelihood estimator can be distorted. This is why at each step, only one variable at a time is considered that contributes the most to the increase in homogeneity, which is referred to as the stepwise splitting up approach. Lastly, there is a reconsideration of the created subgroups where subgroups with comparable transition probabilities are merged together into the final states. 
\vspace{0.5cm}

One major drawback of the procedure in \textcite{de2006modelling} is that prior information is used to determine variables that are thought to be significant for transition behavior, which could introduce bias. \textcite{rombaut2016turnover} proposed a different method based on decision trees, which encapsulates all possible explanatory variables and determines split variables and values based on the transitions that can be found in the dataset. The CART algorithm is employed, which maximizes intra-group homogeneity, while maximizing inter-group heterogeneity. 
\vspace{0.5cm}

As using decision trees is an improved method for finding homogeneous subgroups, some limitations can still be pointed out that justify the need for an alternative framework. CART results in groups of convex shape using axis parallel splits, meaning that after one split, only the separate groups are considered anew without considering how they relate to other branches \parencite{wickramarachchi2016hhcart}. In addition, the usage of a stopping criterion to reach a minimum number of group members, such as minimum node size or maximum depth required, introduces a bias, which could alter the output of the groups \parencite{Han2011}.
\vspace{0.5cm}

The importance of finding appropriate states can be extended to other economic applications than manpower planning using Markov models. Within a same field, state construction is often handled differently, reducing the ability to compare results. For example, in customer lifetime value (CLV) research based on Markov models, states are often determined differently, decreasing the comparability of models. Where \textcite{cheng2012customer} use regression analysis for the creation of states, \textcite{haenlein2007model} employ CART, and \textcite{chan2010model} merely consider the classes inherently present due to the research problem as states, e.g. potential customers, first-time buyers and active customers.
\newline Other macro-economic phenomena such as GDP \parencite{le2004space} and income disparities \parencite{kerkouch2024dynamic} can be studied with spatial Markov models, where initial classes are defined through discretization. Final states within each class are built based on varying methodologies, lacking a general approach. In order to deal with these variations, we turn to unsupervised learning, able to find latent groupings in the data that display similar transitioning behavior, contributing to a framework for state construction. 

\subsection{Clustering}
Since existing approaches for determining homogeneous states have some limitations, we try to overcome these issues through a new grouping methodology using unsupervised clustering: hierarchically density based clustering (HDBSCAN), spectral clustering or Kohonen self-organizing maps (SOM). 
\vspace{0.5cm}

In general, clustering is an unsupervised learning method that looks for patterns and groupings without a labeled outcome variable. This can be achieved through partitioning the set of data objects into $k$ groups, based on the calculation of the inter-group averages, as described in the K-means problem \parencite{macqueen1967multivariate}. Alternatively, groups can be constructed hierarchically based on the closest distance or linkage between observations in order to maximize within-group similarity and between-group dissimilarity \parencite{nielsen2016hierarchical}. In addition, model-based clustering has been designed, for example based on expectation maximization, assuming clusters exist that have the same probability distribution \parencite{bouveyron2014model}.
With these methods, groups are of convex or spherical shape \parencite{king2014cluster}, and assumptions are made about the underlying density distribution that are not always representative for the dataset. In addition, a desired number of clusters often has to be described that is not necessarily equal to the intrinsic division in the data \parencite{campello2020density}. 
\vspace{0.5cm}

\subsubsection*{Hierarchical Density-based Clustering}
With density-based clustering, these drawbacks are not present, as it looks for regions with high density, containing a high amount of similar observations, which can be distinguished from regions with low density, considered as noise. A well-known clustering algorithm is DBSCAN, in which clusters are built around core points. DBSCAN uses two input parameters to determine core points: the radius that is investigated around objects, and the number of observations that have to be within this radius in order to classify as a core point \parencite{campello2020density}. Observations that are connected to the core are added to the cluster, while others are considered as noise \parencite{bhattacharjee2021survey}. This analysis is performed on a similarity matrix, describing the distance between observations based on all relevant variables. 
\newline Advantages of this method are that it does not require a predetermined number of clusters, and it results in clusters of arbitrary shape, nor does it make any assumptions about the distribution density of observations \parencite{campello2020density}. It outperforms other methods in handling noise in a robust manner \parencite{hanafi2022fast}. However, using a fixed radius introduces the drawback that it cannot analyze clusters of varying densities \parencite{stewart2022implementation}. Moreover, it returns a flat clustering solution, unable to distinguish clusters of various densities nested in each other \parencite{campello2015hierarchical}.
\vspace{0.5cm}

To overcome this issue, we turn to HDBSCAN for our analysis, which performs a hierarchical DBSCAN. The algorithm forms clusters hierarchically and refines them based on their stability \parencite{campello2015hierarchical}. Thus, the researcher only has to specify the minimum number of objects (\textit{minPts}) within a cluster. The optimal number can be determined through sensitivity analysis as different \textit{minPts} could showcase better clustering results \parencite{ghosh2024unsupervised, neto2019efficient}. The only drawback of HDBSCAN is that it is a more complex clustering technique, increasing calculation time \parencite{bhattacharjee2021survey}. 
\vspace{0.5cm}

Based on \textit{minPts}, the HDBSCAN algorithm first calculates the core distance for each observation, which is the minimal radius that contains \textit{minPts} observations. Based on the core distances for each observation, the mutual reachability distance is calculated between all pairwise observations \(x\) and \(y\). This is equal to the maximum of either the core distance of \(x\) or \(y\), or the actual distance between \(x\) and \(y\). A weighted graph is constructed based on the combination of all mutual reachability distances. 
\newline Initially, all observations in the graph are considered to exist within the same cluster. Next, connections between observations are cut and the division gets reevaluated to see if the splits still follow the requirement \textit{minPts}. If they don't comply, all observations within this split are considered as noise. The largest distances get cut first, followed by the smaller ones. This approach is repeated until all observations are labeled as noise. Lastly, the step at which this process reaches the highest stability for the clusters is chosen as division for the optimal clustering solution. 
\vspace{0.5cm}

\subsubsection*{Spectral Clustering}
Another method without any of the previously mentioned drawbacks is spectral clustering \parencite{von2007tutorial}. Rather than focusing on regions with high local density, the method emphasizes the connectivity between observations. A graph is constructed from a similarity matrix \(W\), where nodes represent observations and edge weights represent pairwise similarities between them. Based on this graph, the graph Laplacian is computed as \(L=D-W\), where \(D\) denotes the degree matrix containing the summed connectivity of each observation on its diagonal.
\vspace{0.5cm}

The eigenvalues and corresponding eigenvectors of the graph Laplacian are then analyzed \parencite{von2008consistency}. By arranging the eigenvalues in ascending order, large differences between successive eigenvalues can be visually identified. These differences, commonly referred to as eigengaps, provide an indication of the number of clusters required to capture the principal connectivity patterns within the graph. The elbow point in the eigengap plot is chosen as the optimal number of clusters, as the first \(k\) eigenvectors corresponding to the smallest eigenvalues are considered to capture the dominant structural patterns in the dataset. Lastly, K-means is performed on the eigenvectors, seen as coordinates of observations similar to representations found in the Laplacian. 

\subsubsection*{Kohonen Self-Organizing Maps}
To complete the analysis of the presence of latent profiles in datasets, we evaluate SOM's \parencite{kohonen2002self} as a third method of unsupervised grouping. SOM's are unsupervised neural networks that map high-dimensional data onto a smaller grid while preserving topology \parencite{kohonen2013essentials}. Each unit on the grid is characterized by its unique combination, in the form of a vector gathering all input variables given to the network. Its neural network consists of two layers: the input layer and Kohonen layer, represented by a two-dimensional grid structure of interconnected units. The algorithm works by initially assigning a random prototype vector to each unit. Each observation is then fed to the network and assigned to the Best Matching Unit (BMU) based on a predetermined distance measure. After assignment, the vector of the BMU and its neighbors is updated to better represent the new configuration. This is repeated until all observations have been presented. To assure stability, this process is repeated for a predetermined amount of iterations, resulting in a final grid structure. As a result, observations with similar characteristics are mapped onto neighboring regions of the grid, enabling the identification of latent profiles and local structures within datasets.
\vspace{0.5cm}

The main difficulty for the analysis is that the neural network results in an abundant number of units. These units on their own cannot be considered as separate subgroups. That is why an additional step is needed to provide subgroups with latent profiles, based on agglomerative clustering \parencite{goncalves2008unsupervised}. This method merges units bottom-up by aggregating neighboring units that display the smallest change in within-unit difference based on the Gower distances calculated before. The stopping point for this merging process can be determined through analysis of the Ward linkage criterion \parencite{nielsen2016hierarchical}. This criterion measures the increase in within-group variance resulting from merging two regions. Consequently, groups with highly similar prototype vectors and relatively small within-group dispersion are merged first, shown to be effective in prior research \parencite{kiang2001extending}. The optimal number of clusters representing existing topology is determined at the elbow point, indicating the stage beyond which further merging would combine increasingly dissimilar units and result in substantial increases in within-group variance. 

\section{Metholodogy}
The implementation of unsupervised learning techniques within a Markov context requires necessary pre- and post-processing steps to ensure effective and efficient operation of the algorithms as well as the formation of homogeneous states. The following section describes these steps in detail, while a flowchart of the proposed methodology is provided in Figure 1.
\subsection{Pre-processing}
\subsubsection{Class Division}
The first step consists of dividing the dataset according to the classes \(\mathcal{C}=\{C_1, ..., C_n\}\) inherently associated with the research problem. Within each class \(C_i \, (1 \leq i \leq  n)\) of \(\mathcal{C}\) further splits will be made based on the following methodology. 

\subsubsection*{Feature Selection}
In order to adequately construct subgroups within the predefined classes of \(C\), relevant features have to be selected as the performance and accuracy of supervised and unsupervised models is highly dependent on the predictive quality of the underlying dataset. Datasets are often high-dimensional and subject to ``the curse of dimensionality" \parencite{wei2020novel}, whereby the algorithm is unable to distinguish clusters when attributes are abundant. Thus, it is important to decide what features should be retained to construct the models. Feature selection describes the techniques aimed at identifying and eliminating irrelevant variables and redundant predictors that provide little additional predictive information, retaining only the ones most informative for prediction \parencite{cadenas2013feature}. This procedure is beneficial for both the performance of the machine learning algorithm, as well as the computational costs involved \parencite{hancer2020survey}.
\vspace{0.5cm}

A common used practice for feature selection is the usage of random forests, which allows selection based on a target variable. Features are ranked based on permutation-based importance, which quantify the increase in out-of-bag prediction error resulting from randomly permuting the values of a feature. When features cause larger differences in the prediction error, they are considered as being more informative for the target variable and are thus ranked higher. This technique can be tweaked to optimize for low quality datasets with high dimensions \parencite{cadenas2013feature} or based on the underlying goal of interpretation or prediction \parencite{genuer2010variable}. 
\vspace{0.5cm}

For the analysis, the VSURF package created by \textcite{genuer2015vsurf} is employed, which returns two subsets of variables. The first one is for interpretation purposes and can include some redundant variables based on prediction improvement, while this is not the case for the second subset. 
As input parameter, the number of characteristics can be altered to be chosen at each significant split in the tree. When this is lowered, more randomness exists across trees and some relevant variables can be missed. A stability analysis is thus required across different values of this input factor. After relevant variables have been selected based on VSURF, their correlation is analyzed as highly correlated inputs may distort final clustering result and can still be present in the second subset. If two or more variables have a correlation coefficient higher than 70\%, based on the Pearson correlation coefficient \parencite{benesty2009pearson} for numeric variables and the Cramer's V for categorical ones \parencite{telford2020properties}, the variable is chosen that was ranked the highest in terms of predictability for transition behavior, and selected for the third and final feature subset. This procedure is executed for each of the classes \(C_1, ..., C_n\) and results in the feature subset \(\mathcal{F} = \{F_1, ..., F_m\}\) with \(F_i=\{f_{i,1}, ..., f_{i,l}\}\) consisting of all \(l\) informative variables for \(C_i\). 
\vspace{0.5cm}

\subsubsection*{Gower Distance Matrix}
Both spectral clustering and HDBSCAN require a similarity matrix as input for the analysis. This matrix is constructed based on the distance between observations. For datasets consisting of either continuous or categorical data, this is a straightforward procedure as they can be handled in the same way. However, real-life datasets often consist of a mixture of both categorical and continuous data. A way to handle the distances between observations with mixed data points is through the Gower distance, in which continuous variables get normalized \parencite{gower1971general}. One drawback is that categorical variables tend to have a higher weight compared to numerical variables as continuous variables get normalized, while this is not the case for categorical ones. \textcite{liu2024modified} proposed a new method where numerical variables are weighted based on the inter-quartile range, and the categorical are weighted by a scaling vector equal to the ratio of the average of normalized distances of the continuous to the categorical features. The complete technique also determines the relative importance of each feature by determining the weights of the variables when an unfiltered dataset is given as input. If VSURF results in a limited number of variables and categorical characteristics, the impact of advanced feature-weighted distance metrics such as DAFI-Gower is expected to be modest.
\subsubsection*{Dimensionality Reduction}
As mentioned above, using HDBSCAN entails a more complex algorithm, increasing computing time \parencite{bhattacharjee2021survey}. To account for this, a dimensionality reduction will be performed on the Gower distance matrix \parencite{liu2024modified}. Uniform Manifold Approximation and Projection (UMAP) is proposed, which is a technique that is able to store local neighborhood relationships between observations \parencite{kaverinskiy2025scalable}. However, it struggles at preserving global structure, implying that distances and relative positions between well-separated groups may be distorted \parencite{sanchez2023combination}. Hence, the impact of applying UMAP will be analyzed on the final model. Intuitively, it creates a weighted graph based on a given distance matrix gathering distances for all pairs of observations. It then looks for an objective function that preserves all the characteristics of this weighted graph, and seeks for a low dimensional presentation based on optimization \parencite{mcinnes2018umap}. It has been proven to perform well on several clustering techniques, across different datasets \parencite{allaoui2020considerably}. Next to that, it often outperforms alternative dimensionality reduction techniques \parencite{yang2021dimensionality} such as Principal Component Analysis as it does not assume linear relationships between the data \parencite{yang2021dimensionality} and is able to handle large datasets \parencite{mcinnes2018umap}.

\subsection{Unsupervised Learning}
After feature selection and data preparation, the unsupervised learning techniques are applied in order to separate the classes \(\mathcal{C}\) into subgroups \(\mathcal{SG}\). For the comparative analysis in this study, the set \(\mathcal{SG}\) consists of three subsets each relating to a distinct unsupervised learning technique \(u \in \mathcal{U} = \{HD,SC,SOM\}\), resulting in \(\mathcal{SG} = \{SG_{HD}, SG_{SC}, SG_{SOM}\}\) with \(SG_{u} = \{sg_{u,1}, ..., sg_{u,n}\}\) for \(u \in \mathcal{U}\), and \(sg_{u,i}\) the subgroups for class \(C_i\) found by the unsupervised learning technique \(u \in \mathcal{U} = \{HD,SC,SOM\}\). 

\subsection{Post-hoc Subgroup Merging and Validation}
The resulting subgroups \(\mathcal{SG} = \{SG_{HD}, SG_{SC}, SG_{SOM}\}\), derived from clustering or neural networks, are further analyzed based on their transition probabilities, since subgroups with similar transition behavior are not relevant for the final Markov model. For each subset \(SG_u\), separate post-hoc merging is performed. Merging these subgroups reduces the number of parameters to be estimated and may help limit overfitting, thereby improving generalizability. For this purpose, a graph-based merging procedure is applied. Each node in the graph represents an initial cluster, for which the average transition rate is calculated across all observations in that cluster. Edges are created between clusters whose transition rates are not significantly different, based on pairwise Fisher exact tests with Benjamini–Hochberg correction to account for multiple comparisons \parencite{thissen2002quick}. Subgroups connected by an edge are therefore considered statistically similar in terms of transition behavior. Maximal cliques are then identified in this graph, where each clique represents a set of subgroups that are pairwise similar in transition behavior \parencite{newman2018networks}. When cliques overlap, priority is given to the clique with the highest internal cohesion, measured by the lowest average pairwise distance between observations in the original distance matrix. This merging process leads to the final division, the homogeneous states \(\mathcal{S} = \{S_{HD}, S_{SC}, S_{SOM}\}\) with \(S_{u} = \{s_{u,1}, ..., s_{u,n}\}\), \(u \in \mathcal{U}\) and \(s_{u,i}\) the states found for class \(C_i\) after applying the complete three-step procedure. 
\vspace{0.5cm}

The final framework is presented in the flowchart, as displayed below. 
\begin{figure}[ht]
\centering
\resizebox{\textwidth}{!}{%
\begin{tikzpicture}[
    >=Latex,
    thick,
    box/.style={
        draw,
        rounded corners,
        minimum width=3.6cm,
        minimum height=1.2cm,
        align=center,
        font=\small
    }
]
\node[box] (cat) at (0,0)
{Category division $\mathcal{C}$};
\node[box] (vsurf) at (0,-2)
{VSURF variable\\selection};
\node[box] (dist) at (0,-4)
{Gower distance\\ matrix};
\node[box] (umap) at (-4.5,-6)
{UMAP dimensionality\\reduction};
\node[box] (spectral) at (0,-8)
{Spectral clustering\\$(SG_{SC})$};
\node[box] (som) at (4.5,-6)
{SOM};
\node[box] (hdb) at (-4.5,-8)
{HDBSCAN\\$(SG_{HD})$};
\node[box] (agg) at (4.5,-8)
{Agglomerative\\clustering\\$(SG_{SOM})$};
\node[box] (post) at (0,-11)
{Post-hoc subgroup\\merging and validation\\
$\mathcal{S}=\{S_{HD},S_{SC},S_{SOM}\}$};
\node[box] (markov) at (0,-13)
{Markov application\\based on states $\mathcal{S}$};
\draw[->] (cat) -- (vsurf);
\draw[->] (vsurf) -- (dist);
\draw[->] (dist) -- (spectral);
\draw[->] (spectral) -- (post);
\draw[->] (post) -- (markov);
\draw[->] (umap) -- (hdb);
\draw[->] (som) -- (agg);
\draw[->]
(dist.west)
-- ++(0,0)
-| (umap.north);
\draw[->]
(vsurf.east)
-- ++(0,0)
-| (som.north);
\draw[->]
(hdb.south)
-- ++(0,-0.7)
|- (post.west);
\draw[->]
(agg.south)
-- ++(0,-0.7)
|- (post.east);
\draw[dashed,->]
(vsurf.west)
-- ++(-5,0)
node[midway,above,font=\scriptsize,align=center]
{\(\forall\)$C_i\in\mathcal{C}$}
|- (post.west);
\end{tikzpicture}
}

\caption{Flowchart of the proposed clustering and validation procedure.}
\label{fig:method-flowchart}

\end{figure}
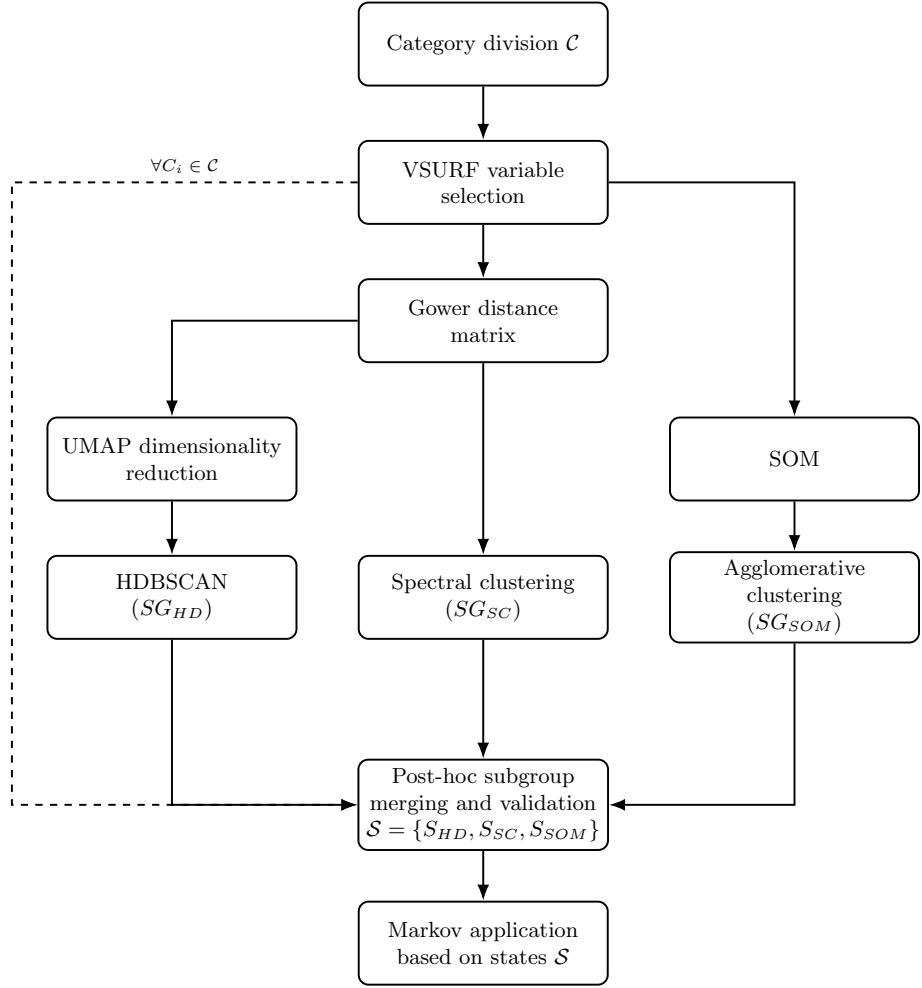

\subsection{Accuracy Measures}
\subsubsection*{Akaike Information Criterion}
A Markov model fits the underlying data when the model-implied covariance matrix matches the empirical covariance matrix. 
\newline The general \(\chi^2\)-test statistic offers insight in the significance of the estimated covariance matrix \parencite{schermelleh2003evaluating}. It tests the hypothesis that the empirical covariance matrix \textbf{S} and the model-implied covariance matrix \(\Sigma(\hat{\theta})\), where \(\hat{\theta}\)
denotes the parameter estimates obtained by maximizing the likelihood function \(L({\theta})\), i.e.: 
\begin{center}
    \(\textbf{S}-\Sigma(\hat{\theta}) = 0\)
\end{center}
A researcher thus seeks to obtain a non-significant \(( p > 5\%)\) \(\chi^2\)-test statistic, implying that the null hypothesis cannot be rejected and indicating a lack of evidence for differences between the covariance matrices \(\textbf{S}\) and \(\Sigma(\hat{\theta})\). This corresponds to a \(\chi^2\)-test statistic that is close to zero and tends to favor more complex models due to an increased number of degrees of freedom, introducing the need for validation measures that penalize for model complexity. 


When a model is controlled for its complexity and an assessment has to be made on what model fits the underlying dataset best, the Akaike Information Criterion (AIC) is useful. 
Originally, the AIC has been introduced as: 
\begin{center}
    \(AIC = -2\ln(L(\hat{\theta}))+2m\)
\end{center}
with \(m\) equal to the number of estimated parameters. The maximum likelihood consists of the optimized model parameters based on the underlying dataset. 
\newline In Markov applications, \textcite{schermelleh2003evaluating} define the AIC as: 
\begin{center}
    
    \(AIC = \chi^2 + 2m\) 

\end{center}
Intuitively, the AIC aims to find the model that best reduces the overall error of a model. A model is subjected to two errors: an approximation error and an estimation error. This is related to the main problem with model selection: the trade-off between the fit and model complexity \parencite{vrieze2012model}. AIC thus compromises both errors and minimizes the trade-off by penalizing a high \(\chi^2\)-test statistic (approximation error) and the number of free parameters (estimation error). AIC states that it can be better to use a simplified model if this is compensated by a reduction of sampling fluctuations as its parameter estimates are less sensitive to random variation in the observed sample. The model that has the lowest value of the AIC, corresponds to the best fitting model. Other researchers in the field \parencite{verbeken2021discrete, rombaut2016turnover}  have applied these accuracy measures in the context of the evaluation of Markov chains as well, indicating the validity of the measure. 
\vspace{0.5cm}

Besides only analysing the raw values of AIC, it is relevant to consider the differences between resulting values. As stated by other researchers \parencite{symonds2011brief}, there exists some uncertainty when choosing the best approximating model based on raw AIC values because it does not specify the statistical importance that can be attached to the difference in AIC values between the first and second best model. \textcite{burnham2004multimodel} have shown that when the difference relative to the AIC value of the best model is between 2 and 7, alternative models should not be dismissed, however these cut-off values are not unambiguous. Therefore, it is advised to consider Akaike weights, which portray the probability for each model being the best model based on a dataset and all candidate models \parencite{wagenmakers2004aic}. 
\newline Weights are calculated by first determining the difference between  \(AIC_i\) of model $i$ and \(AIC_{\min}\), the smallest AIC among the $R$ candidate models
\begin{equation}
\Delta_i = AIC_i - AIC_{\min},
\end{equation}
where
\begin{equation}
AIC_{\min} = \min_{r=1,\ldots,R} AIC_r.
\end{equation}
The relative likelihood of model $i$ is then given by
\begin{equation}
L_i=\exp\left(-\frac{1}{2}\Delta_i\right).
\end{equation}
Finally, the Akaike weight of model $i$ is obtained by normalizing the relative likelihoods over all candidate models:
\begin{equation}
w_i=
\frac{\exp\left(-\frac{1}{2}\Delta_i\right)}
{\sum_{r=1}^{R}\exp\left(-\frac{1}{2}\Delta_r\right)},
\end{equation}
for which
\begin{equation}
\sum_{i=1}^{R} w_i = 1.
\end{equation}

Moreover, one can evaluate the evidence ratio (ER) introduced by \textcite{burnham2002model} that expresses how much more likely the best model is compared to model \(i\). It is calculated as 
\begin{equation}
ER_i=\frac{w_{\mathrm{best}}}{w_i},
\end{equation}



\subsubsection*{Vuong Test}
To further compare two competing, non-nested models, the Vuong statistic \parencite{vuong1989likelihood} is analyzed that can determine whether differences between models are statistically significant while returning which model is closest to an undefined, true model. It does so by evaluating the variance \(\omega_i\) in the case-wise log-likelihood estimations under two candidate models and testing the null hypothesis that both variances are equal \parencite{schneider2020model}, so that 
\begin{align}
H_0 &: E[\omega_*^2]=0,\\
H_1 &: E[\omega_*^2]> 0.
\end{align}
with 
\begin{equation}
\omega_*^2=Var\left[\log\left(
\frac{f_A(\mathbf{x}_i;\hat{\boldsymbol{\Psi}}_A)}
     {f_B(\mathbf{x}_i;\hat{\boldsymbol{\Psi}}_B)}
\right)\right]
\end{equation}
where $\mathbf{x}_i$ denotes the observed response vector for observation $i$, $\hat{\boldsymbol{\Psi}}_A$ and $\hat{\boldsymbol{\Psi}}_B$ represent the maximum likelihood estimates of the parameter vectors for models $A$ and $B$, respectively, and $f_A(\cdot)$ and $f_B(\cdot)$ denote the corresponding probability density functions evaluated at the observed response vector.
If two models are considered distinguishable, the test can be extended to identify the best fitting model, which is the model that is closest to the true model based on the Kullback-Leiber distance \parencite{merkle2016testing}. 

\section{Application}
The proposed unsupervised learning approach is tested on a manpower planning application. The used dataset consists of yearly records of academic university employees, ranging from PhD up to full professor, referred to as their grade. The following 6 grades are considered, in hierarchical order: PhD (Grade 1; $G_1$), post-doc (Grade 2; $G_2$), assistant professor (Grade 3; $G_3$), associate professor (Grade 4; $G_4$), professor (Grade 5; $G_5$) and full professor (Grade 6; $G_6$). These different grades are the classes inherently present in the research problem, i.e. \(C_i = G_i\), and transitioning between grades can be achieved through promotion. Other variables included are \textit{age}, \textit{sex}, \textit{number of children}, \textit{marital status}, \textit{group} (grade specific groups relating to financing or research/teaching duties), \textit{full time equivalent} (\(FTE\)), \textit{seniority} and \textit{statute} (tenured, temporary or indefinite).
\vspace{0.5cm}

Density-based clustering, spectral clustering and SOM's are applied to the manpower database, and its effective grouping is measured through the evaluation of the AIC on the resulting Markov model. They are compared to the benchmark consisting of promotion and stay rates calculated only on the classes, and to the previously constructed tree model \parencite{rombaut2016turnover}. This leads to a final comparison of the following models: 
\begin{itemize}
    \item Class model: Markov model with states defined by the grades $G_i$ under study. 
    \item Tree model: Markov model with states defined by splits found by the decision tree approach introduced in \parencite{rombaut2016turnover}. 
    \item Hierarchical Density  model: Markov model with states resulting from the HDBSCAN approach. Noise is considered as a separate subgroup, following the ideology that not all observations can be classified according to a certain profile. Robustness analysis was performed on the model where noise is assigned to the closest cluster based on the Gower distance matrix, which displayed worse performance. 
    \item Spectral model: Markov model with states resulting from the spectral clustering algorithm. 
    \item SOM model: Markov model with states defined by topology structure, merged through agglomerative clustering. 
\end{itemize}

\subsection{Pre-processing}
The VSURF algorithm is run on each separate class \(C_i=G_i\) with promotion as the target variable. A stability analysis is conducted over different values of \textit{mtry}, the number of characteristics to be chosen at each split. Each \textit{mtry} is considered five times, and only variables that occur in three out of the five runs are selected for further analysis. From this set of variables, the final features subset \(F_i\) consists of the non-correlated variables. For Grade 1, the most important features that should be taken into account based on the variable promotion for the cluster analysis are \textit{full time equivalent}, \textit{age}, \textit{seniority} and \textit{group}. This analysis is performed for each grade separately. For Grade 6, the target variable is wastage instead of promotion, as employees are only able to move through this grade if they leave the university. No relevant variables result from this analysis, indicating that no further splits are considered for Grade 6. 
\vspace{0.5cm}

For each of the feature subsets \(F_i \, (1 \leq i \leq 6)\), the similarity matrix is calculated based on the Gower distance. For HDBSCAN, this is followed by the UMAP dimensionality reduction. The nature of the HDBSCAN algorithm relies on local neighborhood relationships and density estimates \parencite{campello2013density}. When only a precomputed distance matrix is available, many of the computational optimizations cannot be exploited efficiently, depicted by very long run times. Consequently, UMAP is employed as a pre-processing step to obtain a low-dimensional embedding that preserves local neighborhood structure while enabling efficient density-based clustering \parencite{mcinnes2018umap}. This combination is commonly used in practice and substantially reduces the computational burden associated with HDBSCAN \parencite{allaoui2020considerably, kaverinskiy2025scalable}. For visualization purposes, UMAP is used to scale to two dimensions. For spectral clustering, the normal Gower matrix is used, because the algorithm is formulated on a graph representation derived from pairwise similarities \parencite{von2007tutorial}. Moreover, the possible introduced bias resulting from the dimensionality reduction is avoided by skipping this step for spectral clustering. SOM's require the original feature space as input \parencite{kohonen2013essentials}, so the calculation of the similarity matrix is not needed. 

\subsection{Clustering}
\subsubsection*{HDBSCAN}
HDBSCAN was performed on the similarity matrix, reduced to two dimensions through UMAP, both for increased accuracy and visualization. The optimal number of minimum points for a cluster was determined through trial and error based on the following criteria: Silhouette Index (SI) \parencite{ibraimoh2024comparison}, Davies-Bouldin Index \parencite{ros2023pdbi}, the number of clusters, the percentage of noise, mean membership probability and stability. The focus is put on reaching a limited number of clusters in which noise is minimized, e.g. the number of observations that did not get a cluster assignment but are grouped together in cluster 0. Noteworthy is that the SI prefers round-shaped groupings and evenly sized clusters and is not considered very useful for assessing the quality clusters of arbitrary shape \parencite{moulavi2014density}. 
\newline An example is displayed below, for the transition of professor (Grade 5) to full professor (Grade 6), illustrating the UMAP embedding colored according to the HDBSCAN cluster assignments. The visualization suggests the presence of distinct regions in the employee feature space and demonstrates the ability of HDBSCAN to identify non-convex clusters. Because UMAP primarily preserves the local neighborhood structure, the figure should be interpreted as an illustrative representation of the clustering solution rather than an exact depiction of distances in the original feature space.
\begin{figure}[H]
    \centering
    \includegraphics[width=0.7\textwidth]{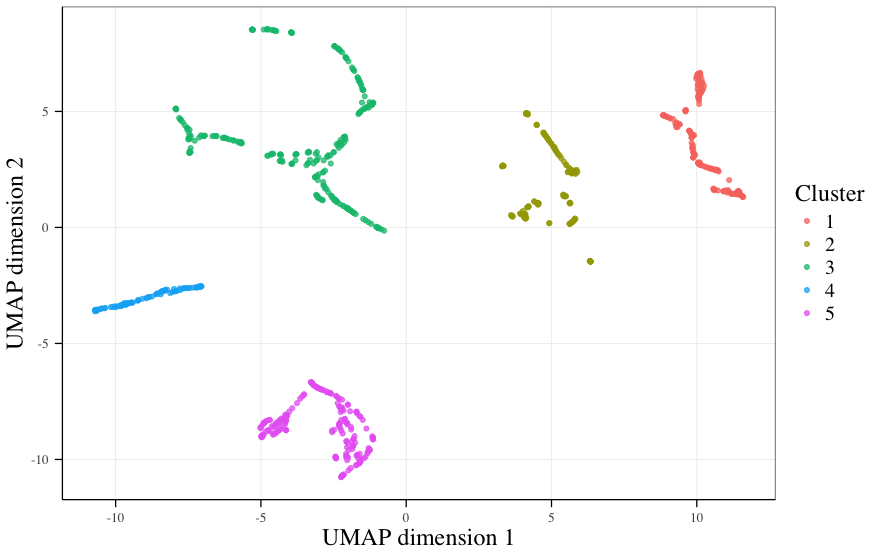}
    \caption{UMAP embedding of Grade 5 employees coloured according to their HDBSCAN cluster assignment.}
    \label{fig:hdbscan_g5}
\end{figure}
\subsubsection*{Spectral Clustering}
For the construction of the spectral subgroups, the eigengaps need to be evaluated. By arranging the eigenvalues from smallest to largest, eigengaps can be identified, which provide an indication of the appropriate number of clusters required to capture the main connectivity patterns within the graph \parencite{von2008consistency}. The cumulative proportion of explained spectral information is displayed as a function of the number of clusters ($k$). An example is provided below, where the elbow point at \(k=4\) suggests that the first four eigenvectors capture the dominant structural patterns in the dataset, as the curve begins to flatten, indicating that the increased information with additional subgroups becomes smaller. The observations are subsequently represented in this spectral space using these eigenvectors and clustered through K-means with \(k=4\). 
\begin{figure}[H]
    \centering
    \includegraphics[width=0.75\textwidth]{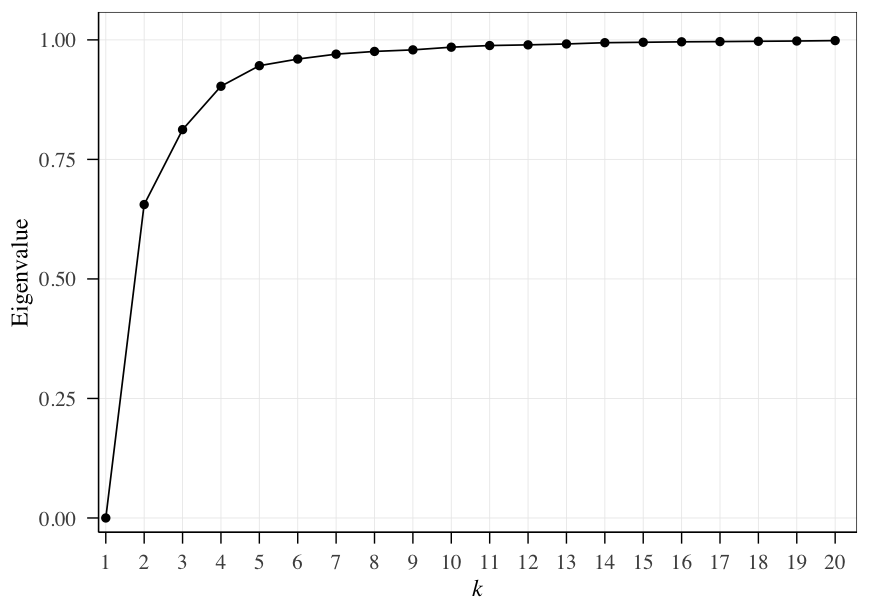}
    \caption{Eigengap plot for the Grade 5 transition dataset.}
    \label{fig:eigengap_g45}
\end{figure}

\subsubsection*{Kohonen Self-Organizing Maps}
For the SOM's, the first step is to determine the number of units in the grid, determined through sensitivity analysis. This is done based on the number of observations per unit and the edges between the nodes. Grid sizes that are able to capture the underlying topology well, display counts that are equally spread across the chart, with localized higher distances. Besides the grid size, the number of iterations is required as input. Larger datasets require a larger amount of training considering it takes more time to converge, which can be measured through the average distance from observations to the closest unit. The learning rate alpha determines how fast prototype vectors move toward observations. Higher alpha leads to less computation time, but noisier maps. 
\newline An example of a diagnostic grid is given below in Figure 4. It clearly shows convergence in plot (a) as the mean distances reaches a plateau, both for the numeric variables as for the categorical variables, displayed by the black and red graph respectively. Moreover, counts are spread evenly in plot (b) showing `counts per SOM unit', and same regions light up as clearly separated through the `neighborhood distances' in plot (c). Plot (d) shows the contribution of the numeric variables to the prototype vectors, with \(FTE\) being the variable name for full time equivalent. The sections within each unit displays the relative value that it contributes to the prototype vector. Categorical variables are left out as they cannot be averaged in the same way as numeric ones. Units in the upper left corner display high levels of the variables \textit{age}, \textit{seniority} and \textit{full time equivalent} indicating these units contain employees with senior, high work percentage employees. Furthermore, neighboring units display similar proportions of contribution, illustrating the topological structure imposed by the SOM.
\begin{figure}[H]
\centering
\begin{subfigure}[t]{0.48\textwidth}
    \centering
    \includegraphics[width=\linewidth]{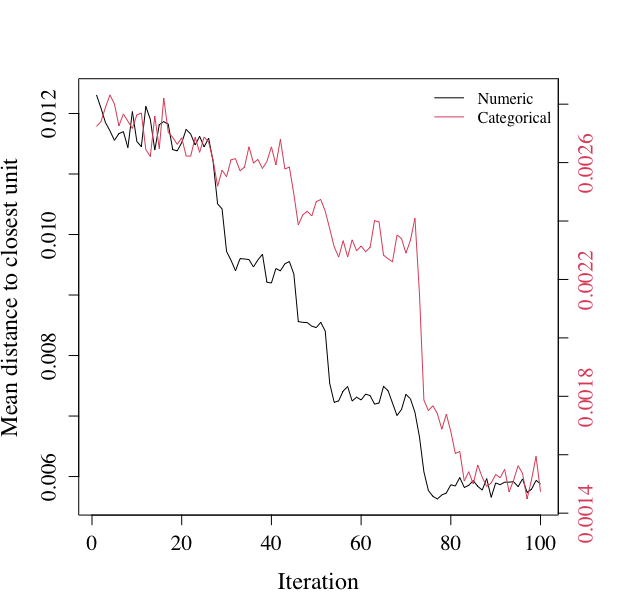}
    \caption{Training progress}
    \label{fig:som-grade5-training}
\end{subfigure}
\hfill
\begin{subfigure}[t]{0.48\textwidth}
    \centering
    \includegraphics[width=\linewidth]{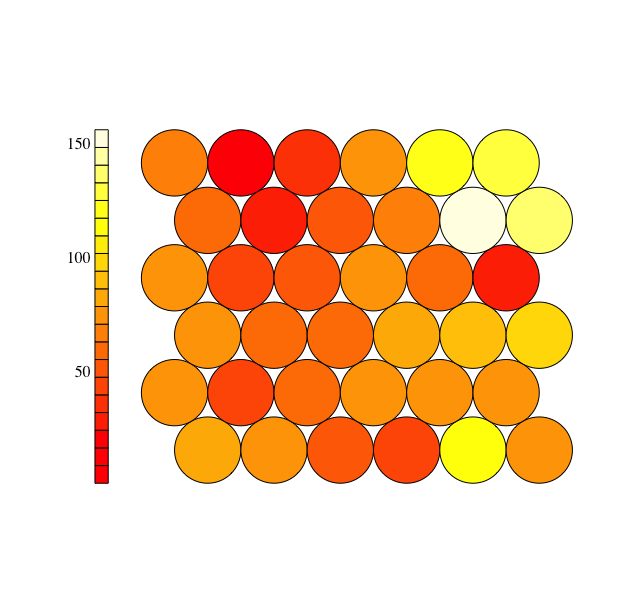}
    \caption{Counts per SOM unit}
    \label{fig:som-grade5-counts}
\end{subfigure}
\vspace{0.4cm}
\begin{subfigure}[t]{0.48\textwidth}
    \centering
    \includegraphics[width=\linewidth]{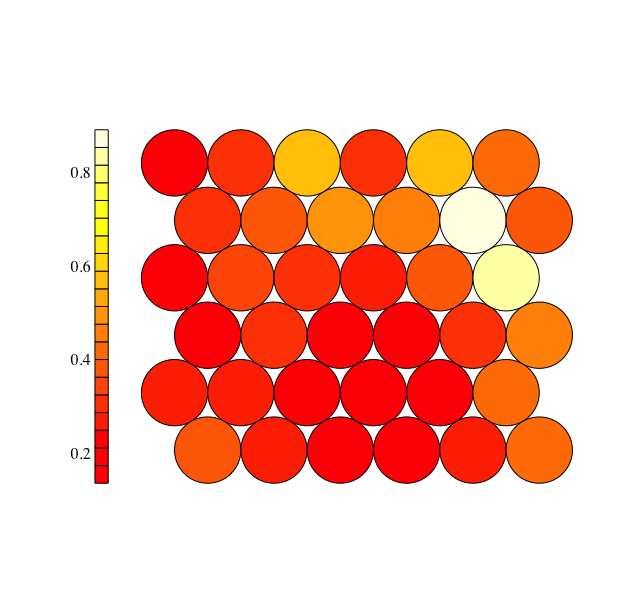}
    \caption{Neighbour distances}
    \label{fig:som-grade5-neighbours}
\end{subfigure}
\hfill
\begin{subfigure}[t]{0.48\textwidth}
    \centering
    \includegraphics[width=\linewidth]{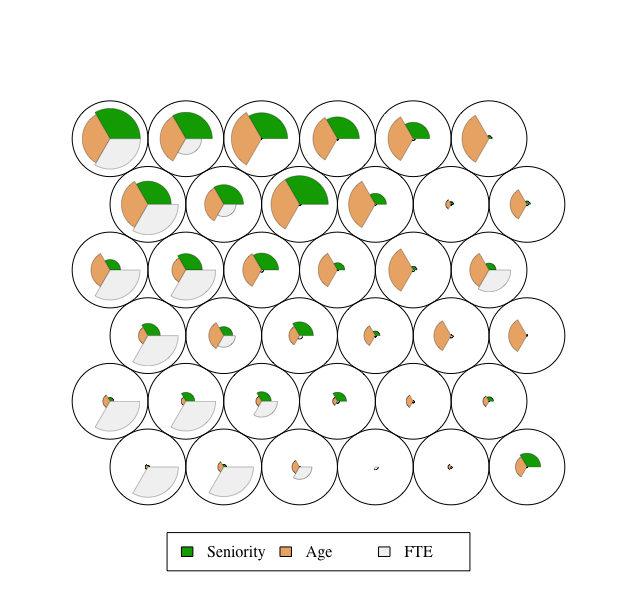}
    \caption{Codebook vectors}
    \label{fig:som-grade5-codebook}
\end{subfigure}
\caption{Diagnostic plots of the self-organizing map (SOM) for Grade~3 employees.}
\label{fig:som-grade5-diagnostics}

\end{figure}

After the number of units is determined, they are merged based on the Ward's linkage criterion. Figure 5 shows the increase in Ward's criterion resulting from successive merges of neighboring SOM units. Large increases indicate the combination of dissimilar subgroups and can be used to identify an appropriate number of final clusters. The optimal number of clusters representing existing topology is determined at the elbow point, indicating in strong within-cluster differences after merging, which can clearly be seen at \(k=4\) in the example below. 
\begin{figure}[H]
    \centering
    \includegraphics[width=0.75\textwidth]{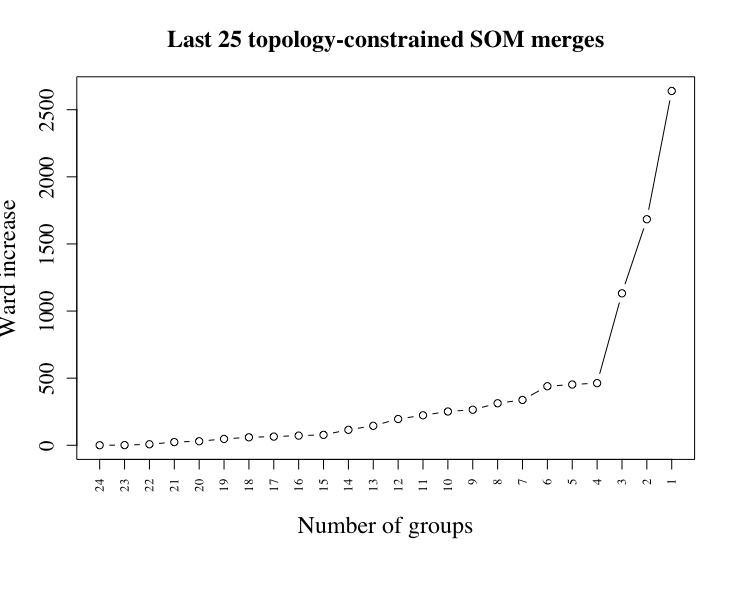}
    \caption{Topology-constrained SOM merge plot for the Grade 2 transition dataset.}
    \label{fig:sommerge_g45}
\end{figure}

\subsection{Post-processing}
In the last step, subgroups obtained from previous methods are merged based on similar transition probabilities through graph-based merging, resulting in the final states used for the Markov model. Nodes consist of the subgroups, and edges represent the alpha value based on the pairwise Fisher test calculated on the transition probabilities between subgroups. The effect of merging subgroups is displayed in the table below. 

\begin{table}[H]
\centering
\caption{Number of subgroups before and resulting states after transition-based merging.}
\label{tab:n_clusters}
\begin{tabular}{lrrrrrrr}
\toprule
& & \multicolumn{2}{c}{HDBSCAN} & \multicolumn{2}{c}{Spectral} & \multicolumn{2}{c}{SOM} \\
\cmidrule(lr){3-4} \cmidrule(lr){5-6} \cmidrule(lr){7-8}
Grade & Tree & Before & After & Before & After & Before & After \\
\midrule
$G_1$ & 4 & 9 & 3 & 2 & 2 & 3 & 2 \\
$G_2$ & 1 & 3 & 2 & 1 & 1 & 4 & 3 \\
$G_3$ & 3 & 3 & 2 & 2 & 2 & 4 & 2 \\
$G_4$ & 2 & 4 & 2 & 3 & 2 & 4 & 2 \\
$G_5$ & 2 & 5 & 3 & 3 & 2 & 3 & 2 \\
\bottomrule
\end{tabular}
\end{table}

\subsection{Results}
Before digging into validation measures, we can observe considerable differences in transition probabilities between the class states and the states found by the alternative approaches. The promotion averages are much more widespread for the unsupervised states, whereas the classes tend to aggregate these deviations. This indicates that classes are heterogeneous, introducing the need to consider subgroups. When comparing across the different techniques, we can see that the spectral states and HDBSCAN states (Figure 6 (b) and (c) respectively) are relatively similar concerning transition probabilities, which are more or less in line with tree promotion rates (Figure 6 (a)). The tree and HDBSCAN states display more states compared to spectral and SOM, of which the probabilities are closer to each other indicating possible overfragmentation. Interestingly, the SOM model displays very different transition probabilities compared to all other methods. Aside from the promotion probabilities, state size and staying probabilities have been calculated and provided in Table 10 in the appendix, revealing that states are sufficiently large. 
\definecolor{darkgreen}{RGB}{0,100,0}
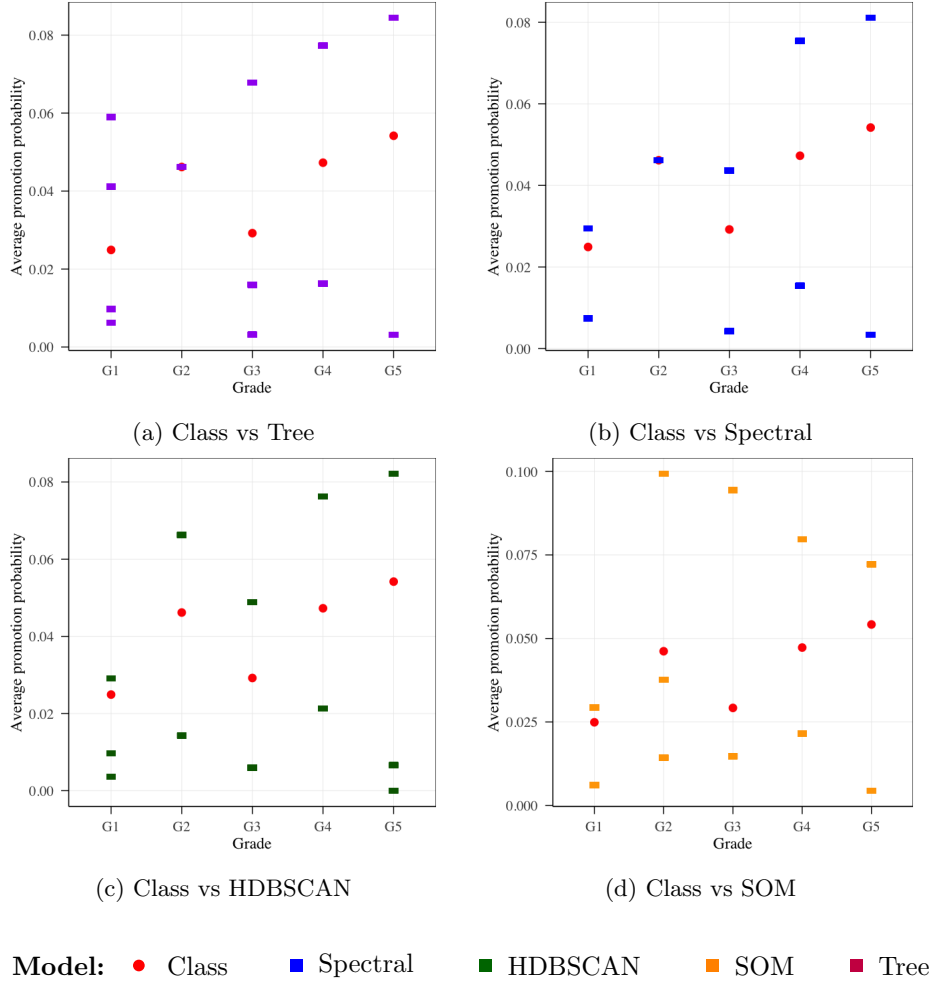
\begin{figure}[H]
    \centering
    \begin{subfigure}{0.48\textwidth}
        \centering
        \includegraphics[width=\linewidth]{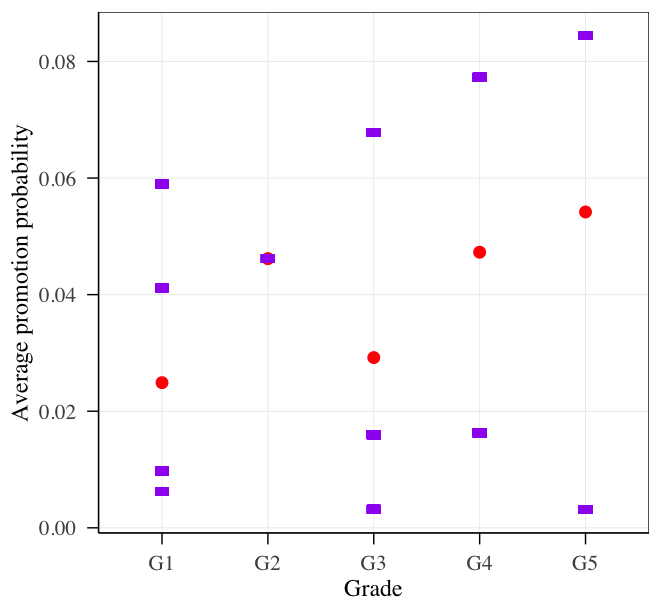}
        \caption{Class vs Tree}
    \end{subfigure}
    \hfill
    \begin{subfigure}{0.48\textwidth}
        \centering
        \includegraphics[width=\linewidth]{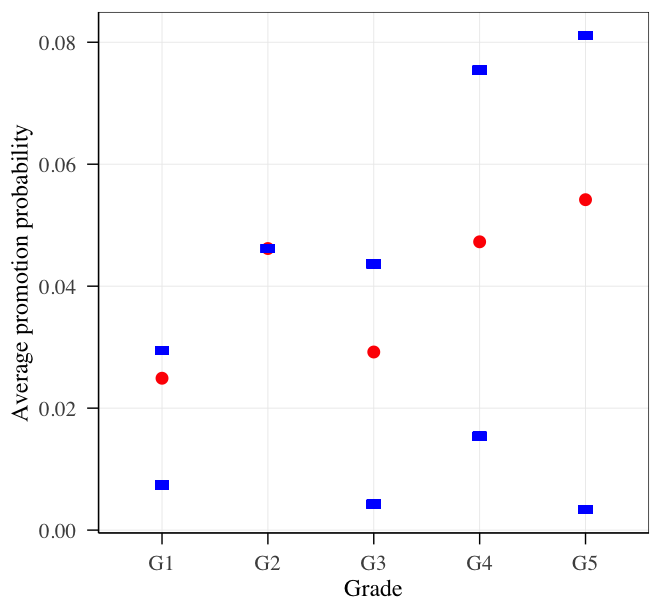}
        \caption{Class vs Spectral}
    \end{subfigure}
    \vspace{0.5cm}
    \begin{subfigure}{0.48\textwidth}
        \centering
        \includegraphics[width=\linewidth]{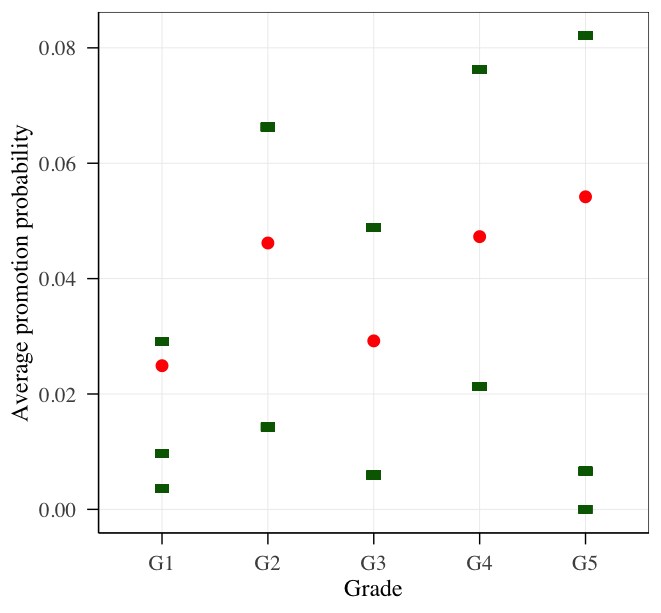}
        \caption{Class vs HDBSCAN}
    \end{subfigure}
    \hfill
    \begin{subfigure}{0.48\textwidth}
        \centering
        \includegraphics[width=\linewidth]{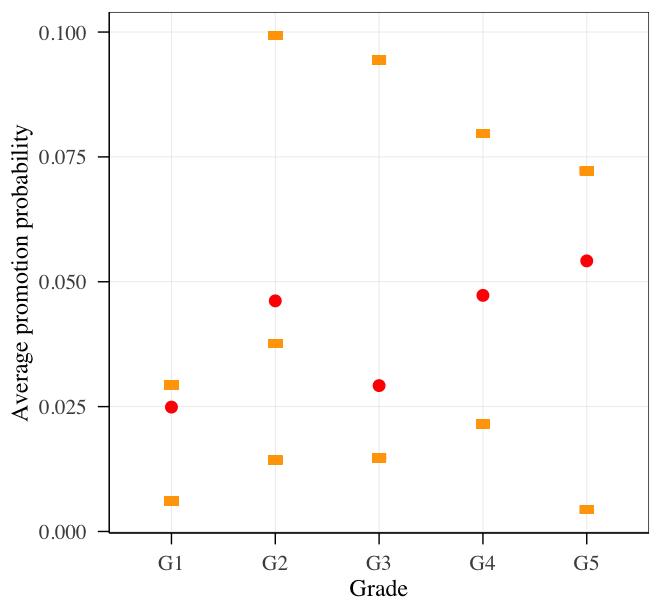}
        \caption{Class vs SOM}
    \end{subfigure}
    \vspace{0.4cm}
    \begin{tikzpicture}
        \node[anchor=west] at (0,0) {\textbf{Model:}};
        \fill[red] (1.8,0) circle (2.2pt);
        \node[anchor=west] at (2.05,0) {Class};
        \fill[blue] (3.8,-0.08) rectangle +(0.16,0.16);
        \node[anchor=west] at (4.05,0) {Spectral};
        \fill[darkgreen] (6.3,-0.08) rectangle +(0.16,0.16);
        \node[anchor=west] at (6.55,0) {HDBSCAN};
        \fill[orange] (9.3,-0.08) rectangle +(0.16,0.16);
        \node[anchor=west] at (9.55,0) {SOM};
        \fill[purple] (11.2,-0.08) rectangle +(0.16,0.16);
        \node[anchor=west] at (11.45,0) {Tree};
    \end{tikzpicture}
    \caption{Transition probabilities for the class model and alternative state construction methods.}
    \label{fig:promotion_comparison}
\end{figure}

A three-fold cross validation gives the following results. In order to respect the longitudinal character of a Markov model, training and test sets are constructed in blocks of three consecutive years. Each fold thus trains on two sets, resulting in six years of training data, and the performance is evaluated on the remaining test set. The states obtained from density-based, spectral clustering and SOM are compared with the class and tree model. 
\newline The models are evaluated based on their \(\chi^2\) statistic, acceptable model fit and ranked on the final AIC. 

\begin{table}[H]
\centering
\small
\setlength{\tabcolsep}{4pt}
\caption{Cross-validated goodness-of-fit measures by grade and model. Values for $\chi^2$ and AIC represent averages across the three folds.}
\label{tab:cv_summary}
\begin{tabular}{llrrrrr}
\toprule
Grade & Model & $\chi^2$ & df & Critical value & Rejected folds & AIC \\
\midrule
$G_1$ & Spectral & 2.31 & 6 & 12.59 & 0 & 14.31 \\
      & SOM      & 2.45 & 10 & 18.31 & 0 & 22.45 \\
      & HDBSCAN  & 7.04 & 15 & 25.00 & 0 & 37.04 \\
      & Tree     & 11.46 & 20 & 31.41 & 0 & 51.46 \\
      & Class    & 54.02 & 2 & 5.99 & 1 & 58.02 \\
\addlinespace
$G_2$ & Spectral & 2.54 & 3 & 7.81 & 0 & 8.54 \\
      & Tree     & 2.57 & 4 & 9.49 & 0 & 10.57 \\
      & Class    & 7.80 & 2 & 5.99 & 1 & 11.80 \\
      & HDBSCAN  & 5.99 & 8 & 15.51 & 0 & 21.99 \\
      & SOM      & 18.38 & 15 & 25.00 & 1 & 48.38 \\
\addlinespace
$G_3$ & SOM      & 1.63 & 8 & 15.51 & 0 & 17.63 \\
      & Class    & 13.64 & 2 & 5.99 & 1 & 17.64 \\
      & Spectral & 2.92 & 8 & 15.51 & 0 & 18.92 \\
      & HDBSCAN  & 3.68 & 8 & 15.51 & 0 & 19.68 \\
      & Tree     & 5.08 & 15 & 25.00 & 0 & 35.08 \\
\addlinespace
$G_4$ & Class    & 4.82 & 2 & 5.99 & 1 & 8.82 \\
      & SOM      & 0.81 & 8 & 15.51 & 0 & 16.81 \\
      & Spectral & 1.44 & 8 & 15.51 & 0 & 17.44 \\
      & Tree     & 2.49 & 8 & 15.51 & 0 & 18.49 \\
      & HDBSCAN  & 2.20 & 10 & 18.31 & 0 & 22.20 \\
\addlinespace
$G_5$ & Class    & 6.87 & 2 & 5.99 & 2 & 10.87 \\
      & Tree     & 3.07 & 6 & 12.59 & 0 & 15.07 \\
      & Spectral & 3.37 & 6 & 12.59 & 0 & 15.37 \\
      & SOM      & 3.98 & 6 & 12.59 & 0 & 15.98 \\
      & HDBSCAN  & 4.33 & 12 & 21.03 & 0 & 28.33 \\
\addlinespace

$G_6$ & HDBSCAN  & 0.19 & 1 & 3.84 & 0 & 2.19 \\
      & SOM      & 0.19 & 1 & 3.84 & 0 & 2.19 \\
      & Spectral & 0.19 & 1 & 3.84 & 0 & 2.19 \\
      & Class    & 0.19 & 1 & 3.84 & 0 & 2.19 \\
      & Tree     & 0.19 & 1 & 3.84 & 0 & 2.19 \\
\bottomrule
\end{tabular}
\end{table}
First, the results highlight the need for disaggregation of the class model. It shows lack of fit through rejected folds, when \(\chi^2\) is larger than the critical value, displaying heterogeneity within these states. The cross validation displays strong and consistent performance of the spectral clustering and SOM models. For Grade 1 and 4, the \(\chi^2\) of spectral and SOM models yield the lowest value compared to the other methods. For Grade 1 and 2, the spectral model outperforms competing models, achieving the lowest AIC, showing that the model fit outweighs the increased complexity. 
Grade 3 has similar AIC for the class and SOM model, however SOM is preferred as \(\chi^2\) is substantially lower indicating a better modeling of transition behavior. For Grade 4, the class model turns out to be the best trade-off between model fit and complexity. However, the rejected goodness-of-fit tests for the class model suggest that heterogeneity remains within these states, implying that the spectral model may provide a more substantively meaningful representation despite its higher complexity. For Grade 5, the smallest class based on amount of observation, the tree model slightly outperforms the spectral model. This might entail that smaller datasets benefit from more aggregated state definitions. The states based on the HDBSCAN model are able to represent flows better than the class states for all grades and the tree states for the majority of the grades, as it yields smaller \(\chi^2\) goodness-of-fit statistic. However, it tends to lead to overfragmentation represented by a high number of states, resulting in many estimated parameters and increased complexity, penalized through the AIC. 
\vspace{0.5cm}

To control for grade specific outcomes, AIC is also compared for each model across all grades and thus for the manpower system as a whole. As \(\chi^2\) is defined as a sum, all grade outcomes displayed prior can be added for each model as long as cells are considered to be independent. This results in the ranking presented in Table 4, showing that the spectral states lead to the best fitting model. The SOM model displays a larger AIC compared to the class model due to a very large \(\chi^2\) value in Grade 2.
\begin{table}[H]
\caption{Overall cross-validated goodness-of-fit measures across all grades.}
\centering
\begin{tabular}[t]{lrrrrr}
\toprule
Model & Total $\chi^2$ & df & Critical value & Rejected folds & Total AIC\\
\midrule
Spectral & 38.33 & 96 & 119.87 & 0 & 230.33\\
Class & 261.99 & 33 & 47.40 & 6 & 327.99\\
SOM & 82.33 & 144 & 173.00 & 1 & 370.33\\
HDBSCAN & 70.29 & 162 & 192.70 & 0 & 394.29\\
Tree & 74.55 & 162 & 192.70 & 0 & 398.55\\
\bottomrule
\end{tabular}
\end{table}

From these total AIC values, corresponding Akaike weights and evidence ratio's are calculated from the averages, as displayed in Table 4. These results portray additional statistical strength for the spectral model, as it got assigned 90\% of the total weights and thus obtaining the strongest support for the candidate models considered. This is followed by the class model, which has to be treated with caution as previous goodness-of-fit tests highlighted rejected folds  during  cross validation for this model. The SOM, HDBSCAN and tree model receive considerably less support. The evidence ratio displays that the spectral model receives approximately 49, 95 and 107 times more support than the candidate models. 
\begin{table}[H]
\caption{Overall Akaike weights across all grades.}
\centering
\begin{tabular}[t]{lrrrr}
\toprule
Model & Mean AIC & \(\Delta\)AIC & Weight & Evidence ratio\\
\midrule
Spectral & 12.80 & 0.00 & 0.90 & 1.00\\
Class & 18.22 & 5.43 & 0.06 & 15.07\\
SOM & 20.57 & 7.78 & 0.02 & 48.86\\
HDBSCAN & 21.90 & 9.11 & 0.01 & 95.04\\
Tree & 22.14 & 9.35 & 0.01 & 107.00\\
\bottomrule
\end{tabular}
\end{table}

Next to the analysis of the Akaike weights, the Vuong test is used to determine whether the differences between the alternative models are statistically significant. The test performs a pairwise comparison between each possible combination of models and analyses which model is preferred if likelihood is distinguishably different. The results are portrayed in Table 5. The class model is eliminated from this analysis, as the prior cross validation already indicated a lack of fit. 
\begin{table}[H]
\caption{Pairwise Vuong tests based on casewise Markov transition log-likelihoods.}
\centering
\begin{tabular}[t]{lrrll}
\toprule
Comparison & \makecell{Mean $\Delta$\\log-likelihood} & $z$ & $p$-value & \makecell{Preferred\\model}\\
\midrule
Spectral vs Tree & 0.214 & 41.801 & 0.00e+00 & Spectral\\
Spectral vs SOM & 0.030 & 7.739 & 9.99e-15 & Spectral\\
Spectral vs HDBSCAN & 0.369 & 20.106 & 0.00e+00 & Spectral\\
Tree vs SOM & -0.184 & -31.900 & 0.00e+00 & SOM\\
Tree vs HDBSCAN & 0.155 & 8.192 & 2.22e-16 & Tree\\
SOM vs HDBSCAN & 0.339 & 18.347 & 0.00e+00 & SOM\\
\bottomrule
\end{tabular}
\end{table}

These statistics reveal interesting insights. First of all, the pairwise comparisons are largely in line with the goodness-of-fit results from the cross validation. The spectral model consistently is preferred over other candidate models, and SOM is preferred over the tree and HDBSCAN model.
\newline However, the Vuong test between the tree and HDBSCAN model should be analysed with caution, as the Vuong test significantly prefers the tree model, whereas the goodness-of-fit based AIC ranks HDBSCAN above the Tree model. This highlights an important difference between the two calculations: prior AIC is calculated based on the goodness-of-fit test, whereas the Vuong test is based on log-likelihood estimations. Tree being the preferred model indicates that observed observations are more likely under the tree model, seen through a higher average log-likelihood contribution across observations. Although the class model achieves a higher likelihood, this does not entail that it is also able to properly construct homogeneous states. 
\newline A possible explanation is that log-likelihood estimations favor high probabilities. These could be more present for the tree model, increasing the favor for this model. Looking back at Table 3, the difference in AIC between the tree and HDBSCAN model is smallest. Consequently, the question can be raised whether focusing on a single model selection criteria is sufficient when differences are small. 
\newline In summary, these results indicate that the spectral model provides an adequate balance between capturing variation through state construction and overall model fit. 

Finally, we consider the HDBSCAN model without the UMAP dimensionality reduction UMAP. The total number of estimated parameters \( (df = 198) \) is higher compared to the states with UMAP \((df = 162)\), which could possibly lead to better estimations. However, the \(\chi^2\) of \(99.90\) is even higher compared to \(74.55\) with UMAP. This shows that the extra divisions do not lead to better estimations, indicating overfragmentation. Thus, the UMAP dimensionality reduction is able to preserve the underlying structure needed for proper Markov modeling. This can be explained through the fact that UMAP only fails to preserve the global structure, while keeping the local structure intact. In our analysis, larger distances between subgroups do not matter for the final construction of states. Tables with cross validation results are provided in appendix.

\section{Discussion and Conclusion}
This study demonstrates how the identification of latent heterogeneity in a dataset helps in constructing homogeneous states for a Markov model. The analysis combines supervised feature selection and clustering algorithms into a three-step grouping methodology to improve homogeneity with respect to transition probabilities. First, the data is split up into classes inherently present in the research problem. Second, classes are separated into subgroups based on unsupervised learning. Lastly, subgroups with equal transition probabilities are aggregated into final states. In contrast to previous supervised learning approaches \parencite{de2006modelling, rombaut2016turnover}, the method can capture complex, non-convex structures and incorporate multiple transition-relevant characteristics simultaneously, which is difficult to represent with consecutive splits based on only variable at a time. 
\newline Empirical results are threefold with a goodness-of-fit statistic, Akaike weights analysis as well as the Vuong test. First off, the AIC goodness-of-fit reveals that heterogeneity remains in the class model, in which states depend on separation inherently present in the research problem, displayed by rejected folds in the cross validation. The overall AIC displays that states derived from spectral clustering outperform the baseline class model, as it reaches the lowest AIC value. All other candidate models, ranked respectively as SOM, HDBSCAN and Tree, outperform the class model based on \(\chi^2\), but are penalized for the higher amount of estimated parameters. Consequently, the improvement in model fit is not influential enough for the increase in model complexity, resulting in higher AIC values compared to the class model, possibly indicating overfragmentation. 
\newline Analysis of the Akaike weights \parencite{wagenmakers2004aic} reveal that the spectral model receives the highest support among all candidate models, with a 90\% probability of being the best approximating model to represent observed transition flows. Its statistical significance is evaluated more with the Vuong test \parencite{vuong1989likelihood}, where it is consistently chosen as the preferred model indicating that it better captures observed transitions. However, some discrepancies exist between the Vuong test and AIC based measures. The class model is preferred over the spectral model, and the tree model is preferred over the HDBSCAN model. This reveals differences in underlying calculations, as well as the importance of including different model selection measures when differences are small. 
\newline The usage of the Vuong test should be argued with its limitations. First off, it does not use degrees of freedom so it is unable to penalize for complexity like AIC based model selection. Besides this, some have \parencite{burnham2011aic} argued against the combination of estimation based and null hypothesis testing because there is no test statistic for the model underneath the null. However, as the Vuong test only considers distinguishable differences between model estimations, this argument does not hold, making the Vuong test a potential interesting additional test for Markov model selection. For the used application, the Vuong test is able to further validate the significance of the spectral model.
\newline Interestingly, the results indicate another possible approach. As there is not one model grouping that consistently seems to outperform alternative groupings when the cross validation is split up over each grade, a hybrid approach can be argued. In this hybrid model, groupings for each grade are determined based on the highest ranked grouping without rejected folds from the cross validation AIC values. This technique is in line with prior research from \textcite{verbeken2021discrete}, where classes are treated differently because each class tends to display distinctive characteristics that benefit from other methods. Aggregated values are displayed in Table 6, and show an AIC that is lower compared to all candidate models. For additional robustness, the Vuong test is applied to compare the spectral model with the hybrid model, displaying significant differences and a preference for the hybrid model, represented in Table 7. 
\begin{table}[H]
\caption{Overall cross-validated goodness-of-fit measures for the hybrid model.}
\centering
\begin{tabular}[t]{lrrrrrr}
\toprule
Model & Total $\chi^2$ &  df & Critical value & Rejected folds & Total AIC \\
\midrule
Hybrid & 31.66  & 96 & 119.87 & 0 & 223.66 \\
\bottomrule
\end{tabular}
\end{table}

\begin{table}[H]
\caption{Overall Vuong test comparing the spectral and hybrid Markov models.}
\centering
\begin{tabular}[t]{lrrll}
\toprule
Comparison & \makecell{Mean $\Delta$\\log-likelihood} & $z$ & $p$-value & \makecell{Preferred\\model}\\
\midrule
Spectral vs Hybrid & -0.016 & -6.87 & 6.44e-12 & Hybrid\\
\bottomrule
\end{tabular}
\end{table}
For the present case study this suggests that employees with comparable characteristics tend to display similar transition behavior, indicating that exploiting these latent similarities can improve state definitions. Through focusing on connectivity, spectral clustering is generally able to reveal states that contribute to the accuracy of the final Markov model displayed by an improved trade-off between model fit and complexity based on the AIC. The framework does require additional testing on alternative datasets in order to further validate its ability to create states, as dataset-specific characteristics of the current application might influence results, reflected through a very large \(\chi^2\) value in Grade 2 for the SOM model.
\newline This research contributes to previous research on state construction by introducing the strength of unsupervised learning to identify latent states displaying similar transition probabilities, which can contribute to enhanced state construction. As states do not arise from predefined splitting criteria or assumptions, the risk of specification error is reduced, leading to improved accuracy and predictive power. 
\newline However, these gains in predictive performance come at the cost of increased implementation complexity and computational effort. Whereas decision trees offer states that are easier to construct and interpret, unsupervised techniques require additional pre- and post processing in order to ensure homogeneity regarding transition probabilities. Practitioners should therefore not only focus on predictive performance, but also consider ease of use. 
\newline This research is subject to some limitations. First, it has been tested only on one application of a first-order Markov model, introducing the need for additional validation. Next up, the hyperparameters for each independent clustering method have not been tested for optimization based on the AIC. Consequently, current state definitions can be suboptimal. However, as the methods outperform previous attempts based on current state definition, optimization could only lead to improved results. Third, the methodology is restricted to discrete-time, time-homogeneous Markov chains and assumes that transition probabilities remain constant over time. 
\newline In conclusion, this research displays how unsupervised learning contributes to state construction for Markov models and shows promising results for improved Markov modeling. 
\section{Further Research}
This study provides a foundation for further research on the integration of machine learning techniques within Markov modeling. Further research can focus on an optimization framework able to incorporate input parameters, pre-and post processing decisions for unsupervised methods resulting in the optimal state construction for the Markov model in terms of model fit and complexity. Pre-processing decisions include alternative mixed-data distance measures, such as the DAFI gower distance or scaling to more than two dimensions with UMAP. Moreover, some applications do not have an inherent class division such as customer lifetime value analyses \parencite{cheng2012customer, haenlein2007model, chan2010model}. For these specific use cases, it could be relevant to asses to what extent the framework helps in defining class division.  In addition to unsupervised learning approaches, future work could explore supervised learning techniques that classify observations based on transition behavior, such as supervised neural networks and support vector machines. Particular attention should be paid to balancing predictive performance, overfitting risk, and interpretability, as highly complex models may be less suitable in practical settings.
\newpage
\section{Generative AI disclosure}
During the preparation of this work the authors used OpenAI ChatGPT (GPT -5.5) in order to generate and debug code, both in R and LaTeX. After using ChatGPT, the authors reviewed and edited the code as needed and take full responsibility for the accuracy and integrity of the code and the results.
\newpage
\printbibliography

@article{suzuki1995markov,
  title={A {M}arkov chain analysis on simple genetic algorithms},
  author={Suzuki, Joe},
  journal={IEEE Transactions on Systems, Man, and Cybernetics},
  volume={25},
  number={4},
  pages={655--659},
  year={1995},
  publisher={IEEE}
}

@article{mcclean1991manpower,
  title={Manpower planning models and their estimation},
  author={McClean, Sally},
  journal={European Journal of Operational Research},
  volume={51},
  number={2},
  pages={179--187},
  year={1991},
  publisher={Elsevier}
}

@article{manasseh2022application,
  title={Application of {M}arkov chain to share price movement in Nigeria (1985--2019)},
  author={Manasseh, Charles O and Iroha, Nnah M and Okere, Kingsley I and Nwakoby, Ifeoma C and Okanya, Ogochukwu C and Nwonye, Nnenna and Odidi, Onuselogu and Inyiama, Oliver I},
  journal={Future Business Journal},
  volume={8},
  number={1},
  pages={59},
  year={2022},
  publisher={Springer}
}

@article{kerkouch2024dynamic,
  title={Dynamic analysis of income disparities in Africa: Spatial Markov chains approach},
  author={Kerkouch, Abderrahim and Bensbahou, Aziz and Seyagh, Intissar and Agouram, Jamal},
  journal={Scientific African},
  volume={24},
  pages={e02236},
  year={2024},
  publisher={Elsevier}
}

@article{bartholomew1975errors,
  title={Errors of prediction for {M}arkov chain models},
  author={Bartholomew, David J},
  journal={Journal of the Royal Statistical Society: Series B (Methodological)},
  volume={37},
  number={3},
  pages={444--456},
  year={1975},
  publisher={Wiley Online Library}
}

@article{de2006modelling,
  title={Modelling heterogeneity in manpower planning: dividing the personnel system into more homogeneous subgroups},
  author={De Feyter, Tim},
  journal={Applied stochastic models in business and industry},
  volume={22},
  number={4},
  pages={321--334},
  year={2006},
  publisher={Wiley Online Library}
}

@inproceedings{rombaut2016turnover,
  title={Turnover Analysis in Manpower Planning},
  author={Rombaut, Evy and Guerry, Marie},
  booktitle={Stochastic Modeling Techniques and Data Analysis International Conference},
  pages={75--76},
  year={2016},
  organization={ISAST: International Society for the Advancement of Science and Technology.}
}

@article{genuer2015vsurf,
  title={VSURF: an R package for variable selection using random forests},
  author={Genuer, Robin and Poggi, Jean-Michel and Tuleau-Malot, Christine},
  journal={The R journal},
  volume={7},
  number={2},
  pages={19--33},
  year={2015}
}

@misc{hastie2009elements,
  title={The elements of statistical learning},
  author={Hastie, Trevor and Tibshirani, Robert and Friedman, Jerome and others},
  year={2009},
  publisher={Springer series in statistics New-York}
}

@inproceedings{campello2013density,
  title={Density-based clustering based on hierarchical density estimates},
  author={Campello, Ricardo JGB and Moulavi, Davoud and Sander, J{\"o}rg},
  booktitle={Pacific-Asia conference on knowledge discovery and data mining},
  pages={160--172},
  year={2013},
  organization={Springer}
}

@incollection{benesty2009pearson,
  title={Pearson correlation coefficient},
  author={Benesty, Jacob and Chen, Jingdong and Huang, Yiteng and Cohen, Israel},
  booktitle={Noise reduction in speech processing},
  pages={1--4},
  year={2009},
  publisher={Springer}
}

@inproceedings{moulavi2014density,
  title={Density-based clustering validation},
  author={Moulavi, Davoud and Jaskowiak, Pablo A and Campello, Ricardo JGB and Zimek, Arthur and Sander, J{\"o}rg},
  booktitle={Proceedings of the 2014 SIAM international conference on data mining},
  pages={839--847},
  year={2014},
  organization={SIAM}
}

@article{von2007tutorial,
  title={A tutorial on spectral clustering},
  author={Von Luxburg, Ulrike},
  journal={Statistics and computing},
  volume={17},
  number={4},
  pages={395--416},
  year={2007},
  publisher={Springer}
}

@article{kohonen2002self,
  title={The self-organizing map},
  author={Kohonen, Teuvo},
  journal={Proceedings of the IEEE},
  volume={78},
  number={9},
  pages={1464--1480},
  year={2002},
  publisher={IEEE}
}

@book{norris1998markov,
  title={Markov chains},
  author={Norris, James R},
  number={2},
  year={1998},
  publisher={Cambridge university press}
}

@book{bremaud2013markov,
  title={Markov chains: Gibbs fields, Monte Carlo simulation, and queues},
  author={Br{\'e}maud, Pierre},
  volume={31},
  year={2013},
  publisher={Springer Science \& Business Media}
}

@book{sericola2013markov,
  title={Markov chains: theory and applications},
  author={Sericola, Bruno},
  year={2013},
  publisher={John Wiley \& Sons}
}

@article{bakhtiari2026prediction,
  title={Prediction of consumer credit card risk from an analysis of spending categories: a hidden {M}arkov model},
  author={Bakhtiari, Ali and Murthi, BPS and Steffes, Erin Marshall},
  journal={International Journal of Bank Marketing},
  volume={44},
  number={2},
  pages={334--353},
  year={2026},
  publisher={Emerald Publishing Limited}
}

@article{ugwuowo2000modelling,
  title={Modelling heterogeneity in a manpower system: a review},
  author={Ugwuowo, Fidelis I and McClean, Sally I},
  journal={Applied stochastic models in business and industry},
  volume={16},
  number={2},
  pages={99--110},
  year={2000},
  publisher={Wiley Online Library}
}

@article{wickramarachchi2016hhcart,
  title={HHCART: an oblique decision tree},
  author={Wickramarachchi, Darshana Chitraka and Robertson, Blair Lennon and Reale, Marco and Price, Christopher John and Brown, Jennifer},
  journal={Computational Statistics \& Data Analysis},
  volume={96},
  pages={12--23},
  year={2016},
  publisher={Elsevier}
}

@book{Han2011,
  author    = {Jiawei Han and Micheline Kamber and Jian Pei},
  title     = {Data Mining: Concepts and Techniques},
  series    = {The Morgan Kaufmann Series in Data Management Systems},
  edition   = {3},
  publisher = {Elsevier},
  year      = {2011},
  isbn      = {9780123814807},
  pages     = {744}
}

@article{cheng2012customer,
  title={Customer lifetime value prediction by a {M}arkov chain based data mining model: Application to an auto repair and maintenance company in Taiwan},
  author={Cheng, C-J and Chiu, SW and Cheng, C-B and Wu, J-Y},
  journal={Scientia Iranica},
  volume={19},
  number={3},
  pages={849--855},
  year={2012},
  publisher={Elsevier}
}

@article{haenlein2007model,
  title={A model to determine customer lifetime value in a retail banking context},
  author={Haenlein, Michael and Kaplan, Andreas M and Beeser, Anemone J},
  journal={European Management Journal},
  volume={25},
  number={3},
  pages={221--234},
  year={2007},
  publisher={Elsevier}
}

@article{chan2010model,
  title={A model for predicting customer value from perspectives of product attractiveness and marketing strategy},
  author={Chan, SL and Ip, WH and Cho, Vincent},
  journal={Expert Systems with Applications},
  volume={37},
  number={2},
  pages={1207--1215},
  year={2010},
  publisher={Elsevier}
}

@inproceedings{macqueen1967multivariate,
  title={Multivariate observations},
  author={MacQueen, J},
  booktitle={Proceedings ofthe 5th Berkeley symposium on mathematical statisticsand probability},
  volume={1},
  pages={281--297},
  year={1967},
  organization={University of California press Oakland, CA, USA}
}

@incollection{nielsen2016hierarchical,
  title={Hierarchical clustering},
  author={Nielsen, Frank},
  booktitle={Introduction to HPC with MPI for Data Science},
  pages={195--211},
  year={2016},
  publisher={Springer}
}

@article{bouveyron2014model,
  title={Model-based clustering of high-dimensional data: A review},
  author={Bouveyron, Charles and Brunet-Saumard, Camille},
  journal={Computational Statistics \& Data Analysis},
  volume={71},
  pages={52--78},
  year={2014},
  publisher={Elsevier}
}

@book{king2014cluster,
  title={Cluster analysis and data mining: An introduction},
  author={King, Ronald S},
  year={2014},
  publisher={De Gruyter}
}

@article{campello2020density,
  title={Density-based clustering},
  author={Campello, Ricardo JGB and Kr{\"o}ger, Peer and Sander, J{\"o}rg and Zimek, Arthur},
  journal={Wiley Interdisciplinary Reviews: Data Mining and Knowledge Discovery},
  volume={10},
  number={2},
  pages={e1343},
  year={2020},
  publisher={Wiley Online Library}
}

@article{bhattacharjee2021survey,
  title={A survey of density based clustering algorithms},
  author={Bhattacharjee, Panthadeep and Mitra, Pinaki},
  journal={Frontiers of Computer Science},
  volume={15},
  number={1},
  pages={151308},
  year={2021},
  publisher={Springer}
}

@article{hanafi2022fast,
  title={A fast DBSCAN algorithm for big data based on efficient density calculation},
  author={Hanafi, Nooshin and Saadatfar, Hamid},
  journal={Expert Systems with Applications},
  volume={203},
  pages={117501},
  year={2022},
  publisher={Elsevier}
}

@article{stewart2022implementation,
  title={An implementation of the HDBSCAN* clustering algorithm},
  author={Stewart, Geoffrey and Al-Khassaweneh, Mahmood},
  journal={Applied Sciences},
  volume={12},
  number={5},
  pages={2405},
  year={2022},
  publisher={MDPI}
}

@article{campello2015hierarchical,
  title={Hierarchical density estimates for data clustering, visualization, and outlier detection},
  author={Campello, Ricardo JGB and Moulavi, Davoud and Zimek, Arthur and Sander, J{\"o}rg},
  journal={ACM Transactions on Knowledge Discovery from Data (TKDD)},
  volume={10},
  number={1},
  pages={1--51},
  year={2015},
  publisher={ACM New York, NY, USA}
}

@inproceedings{ghosh2024unsupervised,
  title={Unsupervised Parameter-free Outlier Detection using HDBSCAN* Outlier Profiles},
  author={Ghosh, Kushankur and Naldi, Murilo Coelho and Sander, J{\"o}rg and Choo, Euijin},
  booktitle={2024 IEEE International Conference on Big Data (BigData)},
  pages={7021--7030},
  year={2024},
  organization={IEEE}
}

@article{neto2019efficient,
  title={Efficient computation and visualization of multiple density-based clustering hierarchies},
  author={Neto, Antonio Cavalcante Araujo and Sander, J{\"o}rg and Campello, Ricardo JGB and Nascimento, Mario A},
  journal={IEEE Transactions on Knowledge and Data Engineering},
  volume={33},
  number={8},
  pages={3075--3089},
  year={2019},
  publisher={IEEE}
}

@article{von2008consistency,
  title={Consistency of spectral clustering},
  author={Von Luxburg, Ulrike and Belkin, Mikhail and Bousquet, Olivier},
  journal={The Annals of Statistics},
  pages={555--586},
  year={2008},
  publisher={JSTOR}
}

@article{kohonen2013essentials,
  title={Essentials of the self-organizing map},
  author={Kohonen, Teuvo},
  journal={Neural networks},
  volume={37},
  pages={52--65},
  year={2013},
  publisher={Elsevier}
}

@article{goncalves2008unsupervised,
  title={An unsupervised method of classifying remotely sensed images using Kohonen self-organizing maps and agglomerative hierarchical clustering methods},
  author={Goncalves, Maria Lu{\'\i}sa and Netto, MLA and Costa, Jos{\'e} Alfredo Ferreira and Zullo Junior, J},
  journal={International Journal of Remote Sensing},
  volume={29},
  number={11},
  pages={3171--3207},
  year={2008},
  publisher={Taylor \& Francis}
}

@article{kiang2001extending,
  title={Extending the Kohonen self-organizing map networks for clustering analysis},
  author={Kiang, Melody Y},
  journal={Computational Statistics \& Data Analysis},
  volume={38},
  number={2},
  pages={161--180},
  year={2001},
  publisher={Elsevier}
}

@article{wei2020novel,
  title={A novel hybrid feature selection method based on dynamic feature importance},
  author={Wei, Guangfen and Zhao, Jie and Feng, Yanli and He, Aixiang and Yu, Jun},
  journal={Applied Soft Computing},
  volume={93},
  pages={106337},
  year={2020},
  publisher={Elsevier}
}

@article{cadenas2013feature,
  title={Feature subset selection filter--wrapper based on low quality data},
  author={Cadenas, Jos{\'e} M and Garrido, M Carmen and Mart{\'\i}Nez, Raquel},
  journal={Expert systems with applications},
  volume={40},
  number={16},
  pages={6241--6252},
  year={2013},
  publisher={Elsevier}
}

@article{hancer2020survey,
  title={A survey on feature selection approaches for clustering},
  author={Hancer, Emrah and Xue, Bing and Zhang, Mengjie},
  journal={Artificial intelligence review},
  volume={53},
  number={6},
  pages={4519--4545},
  year={2020},
  publisher={Springer}
}

@article{genuer2010variable,
  title={Variable selection using random forests},
  author={Genuer, Robin and Poggi, Jean-Michel and Tuleau-Malot, Christine},
  journal={Pattern recognition letters},
  volume={31},
  number={14},
  pages={2225--2236},
  year={2010},
  publisher={Elsevier}
}

@article{telford2020properties,
  title={Properties and approximate p-value calculation of the Cramer test},
  author={Telford, Alison and Taylor, Charles C and Wood, Henry M and Gusnanto, Arief},
  journal={Journal of Statistical Computation and Simulation},
  volume={90},
  number={11},
  pages={1965--1981},
  year={2020},
  publisher={Taylor \& Francis}
}

@article{gower1971general,
  title={A general coefficient of similarity and some of its properties},
  author={Gower, John C},
  journal={Biometrics},
  pages={857--871},
  year={1971},
  publisher={JSTOR}
}

@article{liu2024modified,
  title={A modified and weighted {G}ower distance-based clustering analysis for mixed type data: a simulation and empirical analyses},
  author={Liu, Pinyan and Yuan, Han and Ning, Yilin and Chakraborty, Bibhas and Liu, Nan and Peres, Marco Aur{\'e}lio},
  journal={BMC Medical Research Methodology},
  volume={24},
  number={1},
  pages={305},
  year={2024},
  publisher={Springer}
}

@article{kaverinskiy2025scalable,
  title={Scalable clustering of complex ecg health data: Big data clustering analysis with umap and hdbscan},
  author={Kaverinskiy, Vladislav and Chaikovsky, Illya and Mnevets, Anton and Ryzhenko, Tatiana and Bocharov, Mykhailo and Malakhov, Kyrylo},
  journal={Computation},
  volume={13},
  number={6},
  pages={144},
  year={2025},
  publisher={MDPI}
}

@article{sanchez2023combination,
  title={Combination of cluster analysis with dimensionality reduction techniques for pattern recognition studies in healthcare data: Comparing PCA, t-SNE and UMAP},
  author={S{\'a}nchez-Rico, Marina and Hoertel, Nicolas and Alvarado, Jes{\'u}s},
  year={2023},
  publisher={OSF}
}

@article{mcinnes2018umap,
  title={Umap: Uniform manifold approximation and projection for dimension reduction},
  author={McInnes, Leland and Healy, John and Melville, James},
  journal={arXiv preprint arXiv:1802.03426},
  year={2018}
}

@inproceedings{allaoui2020considerably,
  title={Considerably improving clustering algorithms using {UMAP} dimensionality reduction technique: a comparative study},
  author={Allaoui, Mebarka and Kherfi, Mohammed Lamine and Cheriet, Abdelhakim},
  booktitle={International conference on image and signal processing},
  pages={317--325},
  year={2020},
  organization={Springer}
}

@article{yang2021dimensionality,
  title={Dimensionality reduction by {UMAP} reinforces sample heterogeneity analysis in bulk transcriptomic data},
  author={Yang, Yang and Sun, Hongjian and Zhang, Yu and Zhang, Tiefu and Gong, Jialei and Wei, Yunbo and Duan, Yong-Gang and Shu, Minglei and Yang, Yuchen and Wu, Di and others},
  journal={Cell reports},
  volume={36},
  number={4},
  year={2021},
  publisher={Elsevier}
}

@article{thissen2002quick,
  title={Quick and easy implementation of the {B}enjamini-{H}ochberg procedure for controlling the false positive rate in multiple comparisons},
  author={Thissen, David and Steinberg, Lynne and Kuang, Daniel},
  journal={Journal of educational and behavioral statistics},
  volume={27},
  number={1},
  pages={77--83},
  year={2002},
  publisher={Sage Publications Sage CA: Los Angeles, CA}
}

@book{newman2018networks,
  title={Networks},
  author={Newman, Mark},
  year={2018},
  publisher={Oxford university press}
}

@article{schermelleh2003evaluating,
  title={Evaluating the fit of structural equation models: Tests of significance and descriptive goodness-of-fit measures},
  author={Schermelleh-Engel, Karin and Moosbrugger, Helfried and M{\"u}ller, Hans and others},
  journal={Methods of psychological research online},
  volume={8},
  number={2},
  pages={23--74},
  year={2003}
}

@article{vrieze2012model,
  title={Model selection and psychological theory: a discussion of the differences between the Akaike information criterion ({AIC}) and the Bayesian information criterion ({BIC}).},
  author={Vrieze, Scott I},
  journal={Psychological methods},
  volume={17},
  number={2},
  pages={228},
  year={2012},
  publisher={American Psychological Association}
}

@article{verbeken2021discrete,
  title={Discrete time hybrid semi-{M}arkov models in manpower planning},
  author={Verbeken, Brecht and Guerry, Marie-Anne},
  journal={Mathematics},
  volume={9},
  number={14},
  pages={1681},
  year={2021},
  publisher={MDPI}
}

@article{ibraimoh2024comparison,
  title={Comparison of {K}-means and {HDBSCAN} clustering approaches to enhance marketing strategies},
  author={Ibraimoh, RAPHAEL and Aderoba, ADETUNJI},
  journal={no. January},
  year={2024}
}

@article{ros2023pdbi,
  title={{PDBI}: A partitioning {D}avies-{B}ouldin index for clustering evaluation},
  author={Ros, Fr{\'e}d{\'e}ric and Riad, Rabia and Guillaume, Serge},
  journal={Neurocomputing},
  volume={528},
  pages={178--199},
  year={2023},
  publisher={Elsevier}
}

@article{le2004space,
  title={Space-time analysis of {GDP} disparities among {E}uropean regions: A {M}arkov chains approach},
  author={Le Gallo, Julie},
  journal={International Regional Science Review},
  volume={27},
  number={2},
  pages={138--163},
  year={2004},
  publisher={Sage Publications}
}

@book{burnham2002model,
  title={Model selection and multimodel inference: a practical information-theoretic approach},
  author={Burnham, Kenneth P and Anderson, David R},
  year={2002},
  publisher={Springer}
}

@article{wagenmakers2004aic,
  title={{AIC} model selection using {A}kaike weights},
  author={Wagenmakers, Eric-Jan and Farrell, Simon},
  journal={Psychonomic bulletin \& review},
  volume={11},
  number={1},
  pages={192--196},
  year={2004},
  publisher={Springer}
}

@article{symonds2011brief,
  title={A brief guide to model selection, multimodel inference and model averaging in behavioural ecology using {A}kaike’s information criterion},
  author={Symonds, Matthew RE and Moussalli, Adnan},
  journal={Behavioral ecology and sociobiology},
  volume={65},
  number={1},
  pages={13--21},
  year={2011},
  publisher={Springer}
}

@article{burnham2004multimodel,
  title={Multimodel inference: understanding {AIC} and {BIC} in model selection},
  author={Burnham, Kenneth P and Anderson, David R},
  journal={Sociological methods \& research},
  volume={33},
  number={2},
  pages={261--304},
  year={2004},
  publisher={Sage Publications Sage CA: Thousand Oaks, CA}
}

@article{vuong1989likelihood,
  title={Likelihood ratio tests for model selection and non-nested hypotheses},
  author={Vuong, Quang H},
  journal={Econometrica: journal of the Econometric Society},
  pages={307--333},
  year={1989},
  publisher={JSTOR}
}

@article{merkle2016testing,
  title={Testing nonnested structural equation models.},
  author={Merkle, Edgar C and You, Dongjun and Preacher, Kristopher J},
  journal={Psychological Methods},
  volume={21},
  number={2},
  pages={151},
  year={2016},
  publisher={American Psychological Association}
}

@article{schneider2020model,
  title={Model selection of nested and non-nested item response models using {V}uong tests},
  author={Schneider, Lennart and Chalmers, R Philip and Debelak, Rudolf and Merkle, Edgar C},
  journal={Multivariate Behavioral Research},
  volume={55},
  number={5},
  pages={664--684},
  year={2020},
  publisher={Taylor \& Francis}
}

@article{burnham2011aic,
  title={{AIC} model selection and multimodel inference in behavioral ecology: some background, observations, and comparisons},
  author={Burnham, Kenneth P and Anderson, David R and Huyvaert, Kathryn P},
  journal={Behavioral ecology and sociobiology},
  volume={65},
  number={1},
  pages={23--35},
  year={2011},
  publisher={Springer}
}

\newpage
\section*{Appendix}
\begin{table}[H]
\centering
\small
\setlength{\tabcolsep}{4pt}
\caption{Cross-validated goodness-of-fit measures by grade and model (without UMAP). Values for $\chi^2$ and AIC represent averages across the three folds.}
\label{tab:cv_summary (no UMAP)}
\begin{tabular}{llrrrrr}
\toprule
Grade & Model & $\chi^2$ & df & Critical value & Rejected folds & AIC \\
\midrule
$G_1$ & Spectral & 2.31 & 6 & 12.59 & 0 & 14.31 \\
      & SOM      & 2.45 & 10 & 18.31 & 0 & 22.45 \\
      & Tree     & 11.46 & 20 & 31.41 & 0 & 51.46 \\
      & Class    & 54.02 & 2 & 5.99 & 1 & 58.02 \\
      & \shortstack[l]{HDBSCAN\\(no UMAP)} & 13.21 & 28 & 41.34 & 0 & 69.21 \\
\addlinespace
$G_2$ & Spectral & 2.54 & 3 & 7.81 & 0 & 8.54 \\
      & Tree     & 2.57 & 4 & 9.49 & 0 & 10.57 \\
      & Class    & 7.80 & 2 & 5.99 & 1 & 11.80 \\
      & \shortstack[l]{HDBSCAN\\(no UMAP)} & 13.75 & 15 & 25.00 & 0 & 43.75 \\
      & SOM      & 18.38 & 15 & 25.00 & 1 & 48.38 \\
\addlinespace

$G_3$ & SOM      & 1.63 & 8 & 15.51 & 0 & 17.63 \\
      & Class    & 13.64 & 2 & 5.99 & 1 & 17.64 \\
      & \shortstack[l]{HDBSCAN\\(no UMAP)} & 1.85 & 8 & 15.51 & 0 & 17.85 \\
      & Spectral & 2.92 & 8 & 15.51 & 0 & 18.92 \\
      & Tree     & 5.08 & 15 & 25.00 & 0 & 35.08 \\
\addlinespace
$G_4$ & Class    & 4.82 & 2 & 5.99 & 1 & 8.82 \\
      & SOM      & 0.81 & 8 & 15.51 & 0 & 16.81 \\
      & \shortstack[l]{HDBSCAN\\(no UMAP)} & 0.99 & 8 & 15.51 & 0 & 16.99 \\
      & Spectral & 1.44 & 8 & 15.51 & 0 & 17.44 \\
      & Tree     & 2.49 & 8 & 15.51 & 0 & 18.49 \\
\addlinespace
$G_5$ & Class    & 6.87 & 2 & 5.99 & 2 & 10.87 \\
      & Tree     & 3.07 & 6 & 12.59 & 0 & 15.07 \\
      & \shortstack[l]{HDBSCAN\\(no UMAP)} & 3.31 & 6 & 12.59 & 0 & 15.31 \\
      & Spectral & 3.37 & 6 & 12.59 & 0 & 15.37 \\
      & SOM      & 3.98 & 6 & 12.59 & 0 & 15.98 \\
\addlinespace
$G_6$ & \shortstack[l]{HDBSCAN\\(no UMAP)} & 0.19 & 1 & 3.84 & 0 & 2.19 \\
      & SOM      & 0.19 & 1 & 3.84 & 0 & 2.19 \\
      & Spectral & 0.19 & 1 & 3.84 & 0 & 2.19 \\
      & Class    & 0.19 & 1 & 3.84 & 0 & 2.19 \\
      & Tree     & 0.19 & 1 & 3.84 & 0 & 2.19 \\
\bottomrule
\end{tabular}
\end{table}
\begin{table}[H]
\caption{Overall cross-validated goodness-of-fit measures across all grades without UMAP.}
\centering
\begin{tabular}[t]{lrrrrr}
\toprule
Model & Total $\chi^2$ & df & Critical value & Rejected folds & Total AIC\\
\midrule
Spectral & 38.33 & 96 & 119.87 & 0 & 230.33\\
Class & 261.99 & 33 & 47.40 & 6 & 327.99\\
SOM & 82.33 & 144 & 173.00 & 1 & 370.33\\
Tree & 74.55 & 162 & 192.70 & 0 & 398.55\\
\makecell[l]{HDBSCAN\\(no UMAP)} & 99.90 & 198 & 231.83 & 0 & 495.90\\
\bottomrule
\end{tabular}
\end{table}

\begin{longtable}{lllrcc}
\caption{Observed group sizes and stay/promotion probabilities by grade and model.}\\
\toprule
Grade & Model & Group & $n$ & \makecell{Stay\\probability} & \makecell{Promotion\\probability}\\
\midrule
\endfirsthead

\caption[]{Observed group sizes and stay/promotion probabilities by grade and model. \textit{(continued)}}\\
\toprule
Grade & Model & Group & $n$ & \makecell{Stay\\probability} & \makecell{Promotion\\probability}\\
\midrule
\endhead

\bottomrule
\endlastfoot

G1 & Class & 1 & 12300 & 0.830 & 0.025\\
   & HDBSCAN & 1 & 10029 & 0.827 & 0.027\\
   &         & 2 & 793 & 0.758 & 0.010\\
   &         & 3 & 1478 & 0.888 & 0.003\\
   & Hybrid  & 1 & 9732 & 0.830 & 0.028\\
   &         & 2 & 2568 & 0.828 & 0.006\\
   & SOM     & 1 & 9887 & 0.832 & 0.028\\
   &         & 2 & 2413 & 0.818 & 0.005\\
   & Spectral & 1 & 9732 & 0.830 & 0.028\\
   &          & 2 & 2568 & 0.828 & 0.006\\
   & Tree & 1 & 5854 & 0.902 & 0.006\\
   &      & 2 & 1937 & 0.816 & 0.007\\
   &      & 3 & 933 & 0.795 & 0.038\\
   &      & 4 & 3576 & 0.727 & 0.055\\

\addlinespace[0.8em]

G2 & Class & 1 & 2181 & 0.757 & 0.044\\
   & HDBSCAN & 1 & 843 & 0.751 & 0.014\\
   &         & 2 & 1338 & 0.761 & 0.061\\
   & Hybrid & 1 & 2181 & 0.757 & 0.043\\
   & SOM & 1 & 781 & 0.752 & 0.014\\
   &     & 2 & 795 & 0.770 & 0.035\\
   &     & 3 & 605 & 0.747 & 0.091\\
   & Spectral & 1 & 2181 & 0.757 & 0.043\\
   & Tree & 1 & 2181 & 0.757 & 0.043\\

\addlinespace[0.8em]

G3 & Class & 1 & 2239 & 0.895 & 0.030\\
   & HDBSCAN & 1 & 1238 & 0.901 & 0.049\\
   &         & 2 & 1001 & 0.887 & 0.006\\
   & Hybrid & 1 & 393 & 0.883 & 0.099\\
   &        & 2 & 1846 & 0.897 & 0.015\\
   & SOM & 1 & 393 & 0.883 & 0.099\\
   &     & 2 & 1846 & 0.897 & 0.015\\
   & Spectral & 1 & 807 & 0.874 & 0.005\\
   &          & 2 & 1432 & 0.906 & 0.044\\
   & Tree & 1 & 721 & 0.877 & 0.071\\
   &      & 2 & 551 & 0.889 & 0.004\\
   &      & 3 & 967 & 0.911 & 0.014\\

\addlinespace[0.8em]

G4 & Class & 1 & 1128 & 0.895 & 0.046\\
   & HDBSCAN & 1 & 533 & 0.889 & 0.077\\
   &         & 2 & 595 & 0.901 & 0.018\\
   & Hybrid & 1 & 1128 & 0.895 & 0.046\\
   & SOM & 1 & 498 & 0.892 & 0.078\\
   &     & 2 & 630 & 0.898 & 0.021\\
   & Spectral & 1 & 529 & 0.898 & 0.013\\
   &          & 2 & 599 & 0.893 & 0.075\\
   & Tree & 1 & 556 & 0.899 & 0.014\\
   &      & 2 & 572 & 0.892 & 0.077\\

\addlinespace[0.8em]

G5 & Class & 1 & 775 & 0.871 & 0.055\\
   & HDBSCAN & 1 & 500 & 0.880 & 0.084\\
   &         & 2 & 138 & 0.884 & 0.000\\
   &         & 3 & 137 & 0.825 & 0.000\\
   & Hybrid & 1 & 775 & 0.871 & 0.054\\
   & SOM & 1 & 206 & 0.840 & 0.000\\
   &     & 2 & 569 & 0.882 & 0.074\\
   & Spectral & 1 & 507 & 0.874 & 0.083\\
   &          & 2 & 268 & 0.866 & 0.000\\
   & Tree & 1 & 290 & 0.855 & 0.000\\
   &      & 2 & 485 & 0.880 & 0.087\\

\end{longtable}

\end{document}